\documentclass[letterpaper]{article} 
\usepackage{aaai23}  
\usepackage{times}  
\usepackage{helvet}  
\usepackage{courier}  
\usepackage[hyphens]{url}  
\usepackage{graphicx} 
\usepackage{natbib}  
\usepackage{caption} 
\usepackage{algorithm}
\usepackage{algorithmic}

\usepackage{newfloat}
\usepackage{listings}
\DeclareCaptionStyle{ruled}{labelfont=normalfont,labelsep=colon,strut=off} 
\floatstyle{ruled}
\newfloat{listing}{tb}{lst}{}
\floatname{listing}{Listing}
\usepackage{cuted}
\usepackage{amsmath}
\usepackage{booktabs}
\usepackage{subcaption}
\usepackage{amsfonts}
\usepackage[hang,flushmargin]{footmisc}
\usepackage{xcolor}

\usepackage[capitalize]{cleveref}
\Crefname{table}{Table}{Tables}
\Crefname{figure}{Fig}{Figs}

\title{SHUNIT: Style Harmonization for Unpaired Image-to-Image Translation}
\author{
    Seokbeom Song\textsuperscript{\rm 1},
    Suhyeon Lee\textsuperscript{\rm 1},
    Hongje Seong\textsuperscript{\rm 1},
    Kyoungwon Min\textsuperscript{\rm 2}, and
    Euntai Kim\textsuperscript{\rm 1}\thanks{Corresponding author.}
}
\affiliations{
    \textsuperscript{\rm 1}Yonsei University, Seoul, Korea\\
    \textsuperscript{\rm 2}Korea Electronics Technology Institute, Seongnam, Korea\\
    \{lgs5751, hyeon93, hjseong, etkim\}@yonsei.ac.kr, minkw@keti.re.kr
}

\begin{document}

\twocolumn[{%
    \renewcommand\twocolumn[1][]{#1}%
    \maketitle
    \begin{center}
        \centering
        \includegraphics[width=0.9\linewidth]{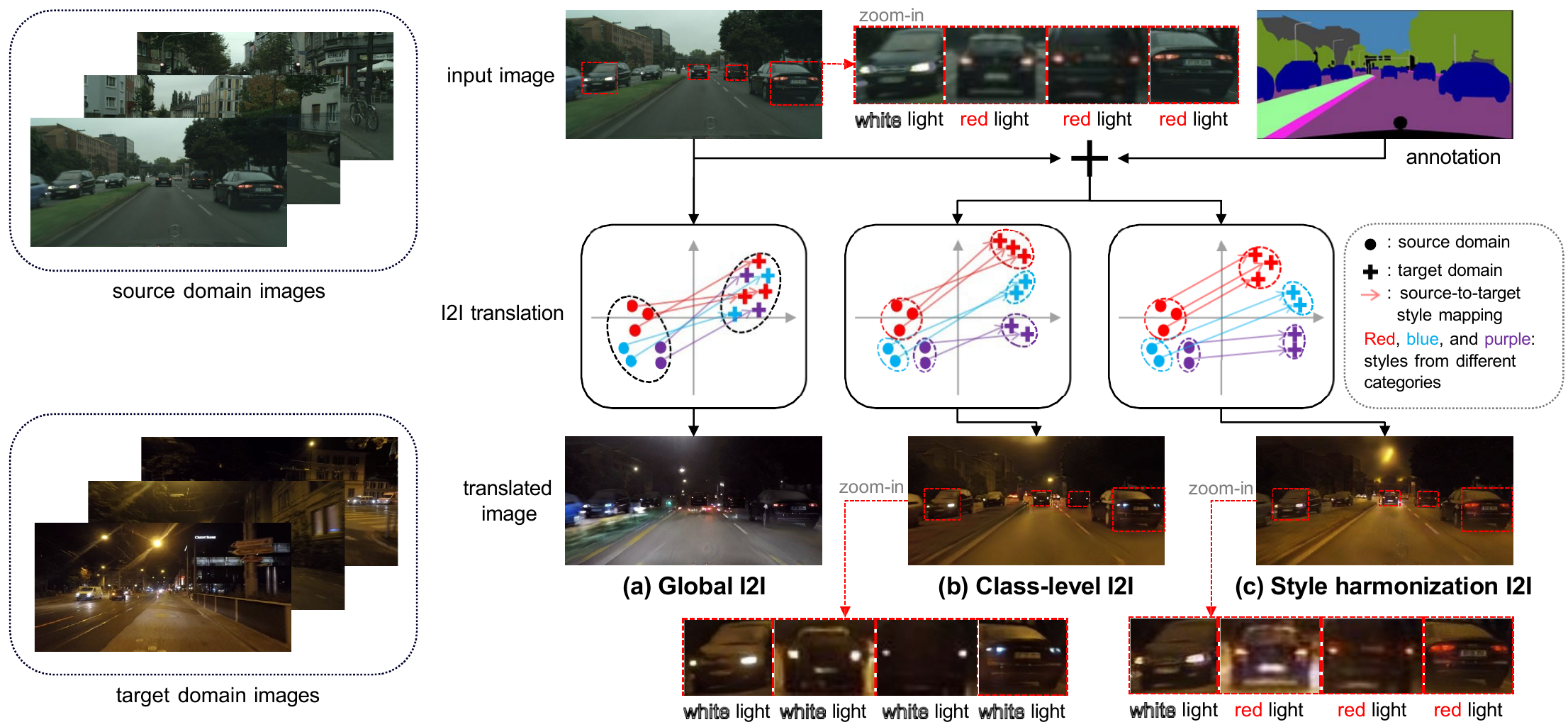}
        \captionof{figure}{Illustration of the concepts in unpaired I2I translation. The results are obtained on Cityscapes~$\rightarrow$~ACDC (night) setting. In the image, many head and tail lamps should be bright at night. (a) \textbf{Global I2I}~\cite{huang2018multimodal} converts all classes to bright because it translates the image with a single source-to-target style mapping function. (b) \textbf{Class-level I2I}~\cite{jeong2021memory} leverages additional annotations to address the problem of (a) and performs per-class source-to-target style mapping. It can effectively deal with multiple classes in an image, but loses the original style: All white lights and red lights become white lights. (c) \textbf{Style harmonization I2I} also performs class-wise style mapping, while adaptively preserving the original styles.}
        \label{fig:concept}
    \end{center}
}]
\renewcommand{\thefootnote}{}
\footnote{\indent *Corresponding authors.}
\renewcommand{\thefootnote}{1}

\begin{abstract} \label{sec:abstract}
We propose a novel solution for unpaired image-to-image (I2I) translation. To translate complex images with a wide range of objects to a different domain, recent approaches often use the object annotations to perform per-class source-to-target style mapping. However, there remains a point for us to exploit in the I2I. An object in each class consists of multiple components, and all the sub-object components have different characteristics. For example, a car in CAR class consists of a car body, tires, windows and head and tail lamps, etc., and they should be handled separately for realistic I2I translation. The simplest solution to the problem will be to use more detailed annotations with sub-object component annotations than the simple object annotations, but it is not possible. The key idea of this paper is to bypass the sub-object component annotations by leveraging the original style of the input image because the original style will include the information about the characteristics of the sub-object components. Specifically, for each pixel, we use not only the per-class style gap between the source and target domains but also the pixel’s original style to determine the target style of a pixel. To this end, we present Style Harmonization for unpaired I2I translation (SHUNIT). Our SHUNIT generates a new style by harmonizing the target domain style retrieved from a class memory and an original source image style. Instead of direct source-\textit{to}-target style mapping, we aim for source \textit{and} target styles harmonization. We validate our method with extensive experiments and achieve state-of-the-art performance on the latest benchmark sets. The source code is available online: {\color{magenta}\url{https://github.com/bluejangbaljang/SHUNIT}}.
\end{abstract}

\section{Introduction} \label{sec:introduction}
Unpaired image-to-image (I2I) translation aims to learn source-to-target style mapping, where source and target images are unpaired.
It can be applied to data augmentation~\cite{antoniou2017data, mariani2018bagan, huang2018auggan, xie2020self}, 
domain adaptation~\cite{hoffman2018cycada, murez2018image} and various image editing applications, such as style transfer~\cite{gatys2016image, huang2017arbitrary, ulyanov2017improved}, colorization~\cite{zhang2016colorful, zhang2017real}, and image inpainting~\cite{iizuka2017globally, pathak2016context}.

In the I2I translation, the biggest problem is how to deal with the style variations among objects or classes.
In other words, when a global style gap is applied to an entire image as in Fig.~\ref{fig:concept}a, the I2I translation often results in unrealistic images because each class has different the style gaps between the source domain and target domain. Recent advanced methods \cite{shen2019towards, bhattacharjee2020dunit, jeong2021memory, kim2022instaformer} addressed the problem by leveraging additional object annotations. They simplify the task into class-level I2I translation and then perform per-class source-to-target style mapping. This enables the networks to explicitly estimate class-wise target styles, but it has a critical limitation: \textit{An object in each class consists of multiple components, and all the sub-object components might also have different characteristics}. Let us consider the example given in Fig.~\ref{fig:concept}b.

Understandably, when a road image taken on a sunny day is translated into a night image, head and tail lamps in a car should be brighter while the rest of the components, such as car body, tires, and windows, should be darker than before. Therefore, each component in a car should be handled separately for realistic I2I translation. However, if the previous approaches are applied to perform per-class source-to-target style mapping, they will translate all head and tail lamps into white lights, making unrealistic images, as shown in Fig.~\ref{fig:concept}b. Here, one might think that this issue can be addressed by annotating more detailed sub-object components than the simple object categories, but it is actually impossible. A brief example is as follows. A car consists of body, window, and tires. A tire consists of wheel and gum. In this way, sub-object components can be divided endlessly.

To solve the above limitation of the previous class-level I2I methods, we present Style Harmonization for unpaired I2I translation (SHUNIT). The key idea of SHUNIT is to bypass the sub-object component annotations by leveraging the original style of the input image because the original style will include the information about the characteristics of the sub-object components. Thus, instead of mapping source-\textit{to}-target style directly, SHUNIT harmonizes the source \textit{and} target styles to realize realistic and practical I2I translation. As illustrated in Fig.~\ref{fig:concept}c, SHUNIT uses not only the per-class style gap between the source and target domains but also the pixel's original style to determine the target style of a pixel. To achieve this, we disentangle the target style into class-aware memory style and image-specific component style. The class-aware memory style is stored in a style memory, and image-specific component style is taken from the original input image.

The goal of the style memory is to obtain class-wise source-to-target style gaps and it is motivated by \cite{jeong2021memory}. Compared to the memory in \cite{jeong2021memory}, our style memory differs in two aspects. First, the output from the style memory was used alone as a target style in \cite{jeong2021memory}, but the output from the memory is adaptively aggregated (=harmonized) in this paper with the style of the original input image to make a target style. Second, the memory was simply updated in \cite{jeong2021memory}, whereas our style memory is jointly trained and optimized with the other parts of SHUNIT. Specifically, the class-aware memory in \cite{jeong2021memory} was not trained but simply was updated using the input features during the training, memorizing the style features from the target domain. The gradient was not propagated to the style memory. Thus, the style memory in \cite{jeong2021memory} cannot update their parameters based on the error of memory. In SHUNIT, however, we overcome this problem by enabling the memory to learn through backpropagation.
To this end, we train the style memory from randomly initialized parameters and introduce style contrastive loss to constrain the memory to learn class-wise style representations. The backpropagation forces the style memory to reduce the final loss jointly and effectively along with the other parts of SHUNIT. To demonstrate the superiority of our SHUNIT, we conduct extensive experiments on the latest benchmark sets and achieve state-of-the-art performance.

Overall, the contributions of our work are summarized as follows: 
\begin{itemize}
    \item We present a novel challenge in I2I translation: an object might have various styles.
    \item We propose a new I2I method, style harmonization, that leverages two distinct styles: class-aware memory style and image-specific component style. To the best of our knowledge, the style harmonization is the first method to estimate the target style in multiple perspectives for unpaired I2I translation.
    \item We achieve new state-of-the-art performance on latest benchmarks and provide extensive experimental results with analysis.
\end{itemize}

\begin{figure*}[t]
	\centering
	\includegraphics[width=.75\linewidth]{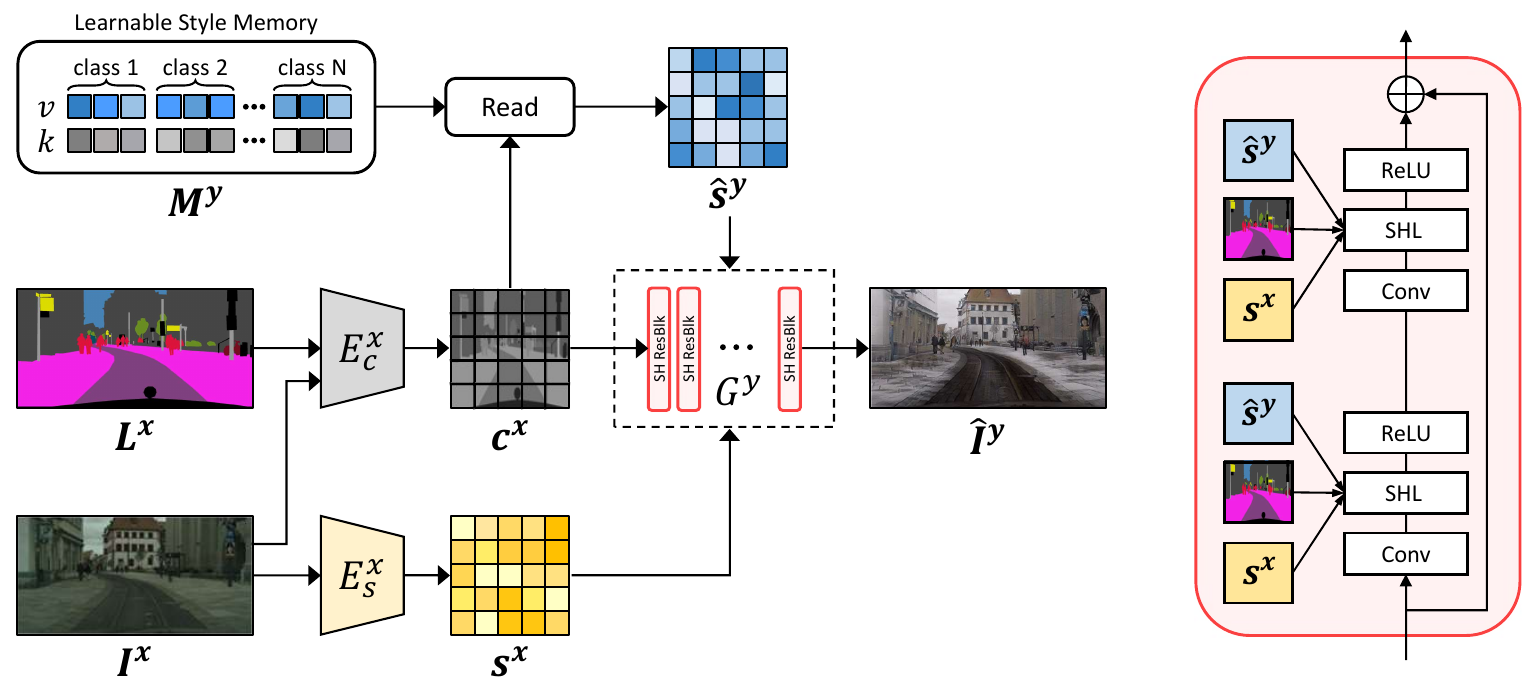}\\
	\raggedright{\small\hspace{150pt} (a) Overall architecture \hspace{132pt} (b) SH ResBlk}
	\caption{\textbf{An overview of SHUNIT.} (a) From a pair of image $I^x$ and label $L^x$ in source domain $\mathcal{X}$, the two encoders (\textit{i.e.}, $E_c^x$ and $E_s^x$) extract the content $c^x$ and component style $s^x$, respectively. The content retrieves the memory style $\hat{s}^y$. The three features $c^x$, $s^x$, and $\hat{s}^y$ are fed to the target generator $G^y$, which consists of several Style Harmonization Residual Blocks (SH ResBlk). The generator $G^y$ outputs the translated image $\hat{I}^y$.
	(b) The input of the SH ResBlk is the content $c^x$ for the first layer, and the output of the previous block is used as input for the remainders. Each SH ResBlk includes two Style Harmonization Layers (SHL) that transfer target styles, \textit{i.e.}, $s^x$ and $\hat{s}^y$.
	}
	\label{fig:architecture}
\end{figure*}

\section{Related Work} \label{sec:related_work}
\paragraph{Image-to-image translation.}
The goal of I2I is to learn source-to-target style mapping.
For I2I translation, pix2pix~\cite{isola2017image} proposes a general solution using conditional generative adversarial networks~\cite{mirza2014conditional}.
However, it has a significant limitation: paired training data should be used for training networks.
CycleGAN~\cite{zhu2017unpaired} successfully addresses this problem with a cycle consistency loss. The loss allows us to train the networks with unpaired training data by supervising the reconstructed original image only.
Based on CycleGAN, many approaches~\cite{ kim2017learning, choi2018stargan} have been proposed to tackle I2I translation take in an unpaired manner.
UNIT~\cite{liu2017unsupervised} proposes another unpaired I2I translation solution by mapping two images in different domains into the same latent code in a shared-latent space.
MUNIT~\cite{huang2018multimodal} and DRIT~\cite{lee2018diverse} introduce a disentangled representation to achieve diverse and multi-modal I2I translation from unpaired data.
Basically, they perform global I2I translation which focuses on mapping a global style on all pixels in an image.
Although they work well on object-centric images, they bring severe artifacts for complex images, such as multiple objects being presented or large domain gap scenarios, as illustrated in Fig.~\ref{fig:concept}a.
To complement the problem, recent approaches leverage additional object annotations and perform class-level I2I translation.

\paragraph{Class-level image-to-image translation.}
Recent several approaches~\cite{mo2018instagan, shen2019towards, bhattacharjee2020dunit, jeong2021memory, kim2022instaformer} propose class-level image-to-image translation solutions with object annotations.
Specifically, INIT~\cite{shen2019towards} generates instance-wise target domain images.
DUNIT~\cite{bhattacharjee2020dunit} additionally employs an object detection network and jointly trains it with the I2I translation network.
MGUIT~\cite{jeong2021memory} proposes an approach to store and read class-wise style representations with key-value memory networks~\cite{miller2016key}.
This approach, however, cannot directly supervise the memory with objective functions for I2I translation.
Instaformer~\cite{kim2022instaformer} proposes a transformer-based \cite{vaswani2017attention,dosovitskiy2020image} architecture that mixes instance-aware content and style representations.
The existing methods that leverage object annotations learns the direct class-wise source-to-target style mapping, as shown in Fig.~\ref{fig:concept}b.
This effectively simplifies the I2I translation problem into per-class I2I translation, but they overlook an important point that not all pixels in the same class should be translated with the same style. Our approach, style harmonization, addresses this problem by introducing the component style that facilitates preserving the original style of the source image, as illustrated in Fig.~\ref{fig:concept}c.

\section{Proposed Method} \label{sec:proposed_method}
\subsection{Definition and Overview}
Let $\mathcal{X}$ and $\mathcal{Y}$ be the visual source and target domains, respectively.
Given an image and the corresponding label (=bounding box or segmentation mask) in $\mathcal{X}$ domain, our framework generates a new image in $\mathcal{Y}$ domain while remaining the semantic information in the given image.
We assume that each domain consists of images and labels denoted by $(I^x, L^x) \in \mathcal{X}$ and $(I^y, L^y) \in \mathcal{Y}$, and both domains have the same set of $N$ classes.
Our framework contains the source encoder $E^x=\{E_{c}^x, E_{s}^x\}$, target generator $G^y$, and target style memory $M^y$ for source-to-target mapping, and the target encoder $E^y=\{E_{c}^y, E_{s}^y\}$, source generator $G^x$, source style memory $M^x$ for target-to-source mapping.
For convenience, we will only describe the source-to-target direction, and the overview of our framework is depicted in Fig.~\ref{fig:architecture}.

Following the previous studies~\cite{huang2018multimodal, lee2018diverse}, we assume that an image can be disentangled into domain-invariant content and domain-specific style. For this, we basically follow the MUNIT~\cite{huang2018multimodal} architecture. 
The content encoder $E_{c}^x$ consists of several strided convolutional layers and residual blocks~\cite{he2016deep}, and all the convolutional layers are followed by Instance Normalization~\cite{ulyanov2016instance}.
The content encoder extracts the domain-invariant \textit{content feature} $c^x$ from the image $I^x$ and label $L^x$.
The style encoder $E_{s}^x$ also consists of several strided convolutional layers and residual blocks, and it extracts the \textit{component style} feature $s^x$ from the image $I^x$.
The \textit{memory style} $\hat{s}^y$ is read by retrieving from the learnable style memory $M^y$ to the content feature $c^x$.
The generator $G^y$ consists of several style harmonization layers and residual blocks, and it produces the translated image $\hat{I}^y$ from $c^x$, $s^x$, and $\hat{s}^y$.

\subsection{Style Harmonization for Unpaired Image-to-Image Translation (SHUNIT)} \label{sec:SHUNIT}
The important point of SHUNIT is that two styles are employed to determine the target style: One is image-specific \textit{component style}, and the other one is class-aware \textit{memory style}.
We focus on extracting two distinct styles accurately and then harmonizing them.
In what follows, we describe the detail of each step.

\paragraph{Component style.}
The style encoder $E_{s}^x$ takes the image $I^x$ as input and extracts the style feature $s^x$ of size ${H}\times{W}\times{C}$, where $H$, $W$, and $C$ are the height, width, and number of channels of the feature, respectively.
Here, $s^x$ explicitly represents the style of the input image, thus we use it as image-specific \textit{component style}.
The component style is used together with the memory style in the style harmonization layer to reduce the artifacts of the generated target image.

\paragraph{Memory style.}
The component style is not sufficient to handle complex scenes with multiple objects.
Therefore, we exploit the class-specific style that leverages an object annotation.
To this end, we construct the class-wise style memory retrieved by the content feature.
The target style memory $M^y$ consists of $N$ class memories to store class-wise style representations of the target domain $\mathcal{Y}$.
Each class memory $M^y_n$ has $U$ key-value pairs $(k^y, v^y)$~\cite{jeong2021memory}, considering that various styles exist in one class (\textit{e.g.}, different styles of headlamp and tires in CAR class).
The key $k^y$ is used for matching with the content feature and the value $v^y$ has the class-aware style representations.
The key and value are learnable vectors, each one of size $1\times1\times{C}$.

\begin{figure}[t]
	\centering
    \includegraphics[width=.85\linewidth]{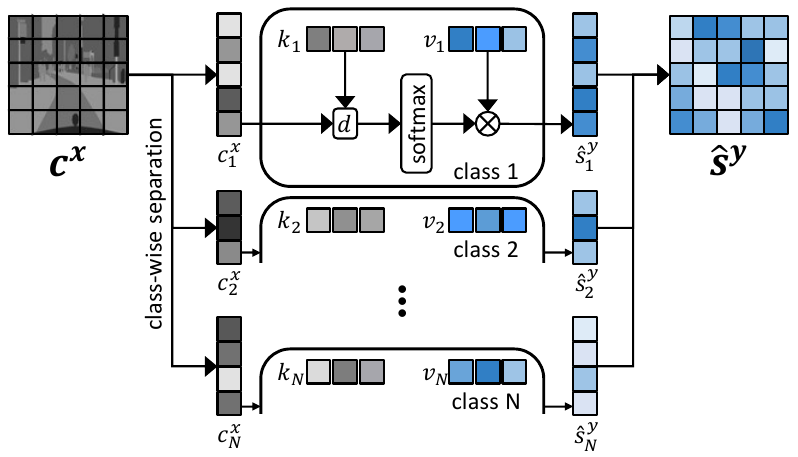}
    \caption{\textbf{Read operation in style memory.} The content feature $c^x$ is separated by the label $L^x$. For each class, the memory read is independently performed. After class-wise memory read, the memory style $\hat{s}^y$ is obtained by gathering the retrieved class-wise values into the original locations. $\bigotimes$ denotes matrix multiplication.}
	\label{fig:read}
\end{figure}

Fig.~\ref{fig:read} shows a detailed implementation of the process of reading the corresponding \textit{memory style} $\hat{s}^y$ from the memory $M^y$.
With the semantic label, we separate the content feature $c^x$ into $\{c_{1}^x, c_{2}^x, \cdots, c_{N}^x\}$, where $c_{n}^x$ denotes source content feature for the $n$-th class.
Let $c_{n, i}^x$ be the $i$-th pixel of the $n$-th class source content feature and $(k_{n, j}^y , v_{n, j}^y)$ be the $j$-th key-value pair of the $n$-th class target style memory $M^y_n$.
In this work, we aim to read the target memory style $\hat{s}_{n,i}^y$ corresponding to the source content $c_{n, i}^x$ using the similarity between the source content and key of target memory.
To this end, we calculate the similarity $w_{n, i, j}^x$ between $c_{n, i}^x$ and $k_{n, j}^{y}$ as:
\begin{equation}
    w_{n, i, j}^x = \frac{\exp \left({d(c_{n, i}^x, k_{n, j}^{y})} \right)}{\sum_{u=1}^U\exp \left({d(c_{n, i}^x, k_{n, u}^{y})} \right)}
\end{equation}
where $d(\cdot,\cdot)$ is the cosine similarity.
We then read the memory style $\hat{s}_{n,i}^y$ corresponding to $c_{n, i}^x$ by calculating the weighted sum of values in $n$-th class style memory:
\begin{equation}
    \hat{s}_{n, i}^y = \sum_{j=1}^{U}w_{n, i, j}^x v_{n, j}^{y} .
\end{equation}
The same process is applied for content features of other classes, finally extracting spatially varying target memory style features $\hat{s}^y$ of size $H \times W \times C$.

Different from the previous key-value memory networks~\cite{jeong2021memory} that learn the memory via updating mechanism, we learn the memory through backpropagation.
The updating mechanism is used to directly store the external input features.
However, it has a critical drawback: The memory cannot be trained with the network jointly with the same objective function because the gradient should be stopped at the updated memory.
To solve the problem, we discard the update mechanism and learn the memory with the loss functions presented in Eq.~(\ref{eq:loss}).
The effectiveness of our memory learning strategy is validated in the experiments section.

\begin{figure}[t]
	\centering
	\includegraphics[width=.65\linewidth]{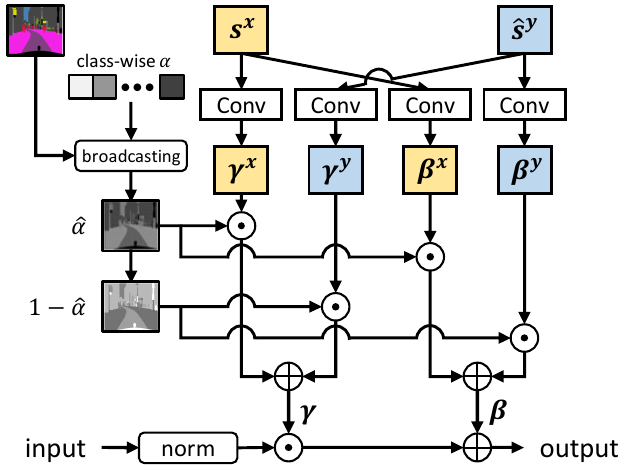}
	\caption{\textbf{Style harmonization layer.} From the component style $s^x$ and memory style $\hat{s}^y$, scale ($\gamma^x$ and $\gamma^y$) and shift ($\beta^x$ and $\beta^y$) factors are extracted via four convolutional layers. The alpha mask $\hat{\alpha}$ is obtained by broadcasting the class-wise alpha $\alpha$ with the semantic label $L^x$. We weighted sum the image and memory styles with the alpha mask, and transfer it to the input feature. $\odot$ denotes Hadamard product.}
	\label{fig:CSFDE}
\end{figure}

\paragraph{Style harmonization layer.}
Our goal is harmonizing the source \textit{and} target styles instead of mapping source-\textit{to}-target style directly.
To this end, we propose the style harmonization layer to adaptively aggregate the component style and memory style.
The style harmonization layer consists of several convolution layers and class-wise alpha parameters, and it is illustrated in Fig.~\ref{fig:CSFDE}.
Here we use three conditional inputs: memory style, component style, and label.
Convolutional layers are used to compute pixel-wise scale $\gamma$ and shift $\beta$ factors from the two styles.
Following~\cite{jiang2020tsit,park2019semantic, zhu2020sean, ling2021region}, we transfer the harmonized target style by scaling and shifting the normalized input feature with the computed factors (\textit{i.e.}, $\gamma$ and $\beta$).
In the layer, we additionally set class-wise alpha parameters $(\alpha_{1}, \alpha_{2}, \cdots, \alpha_{N})$. It is used to decide which style has more influence for each class in the generated images.
If the alpha value is large, the component style $s^x$ has more influence, and vice versa.

Let $f$, of size $H \times W \times C$, be the input feature of the current style harmonization layer in the generator $G^y$.
With the style harmonizing scale $\gamma$ and shift $\beta$ factors, the feature is denormalized by
\begin{equation}
    \gamma_{c,h,w} \frac{(f_{c,h,w} - \mu_c)}{\sigma_c} + \beta_{c,h,w}
\end{equation}
where $\mu_c$ and $\sigma_c$ are the mean and standard deviation of the input feature $f$ at the channel $c$, respectively.
The modulation parameters $\gamma_{c,h,w}$ and $\beta_{c,h,w}$ are obtained from $\gamma_{c,h,w}^x$, $\gamma_{c,h,w}^y$, $\beta_{c,h,w}^x$, and $\beta_{c,h,w}^y$, which are the scale and shift factors of the component style $s^x$ and memory style, $s^y$ respectively, and they are computed by
\begin{align}
\begin{split}
    \gamma_{c,h,w} &= \hat{\alpha}_{h,w}\gamma_{c,h,w}^x + (1-\hat{\alpha}_{h,w})\gamma_{c,h,w}^y, \\
    \beta_{c,h,w} &= \hat{\alpha}_{h,w}\beta_{c,h,w}^x + (1-\hat{\alpha}_{h,w})\beta_{c,h,w}^y,
\end{split}
\end{align}
where $\hat{\alpha}$ denotes the alpha mask. It is obtained by broadcasting the class-wise alpha parameters to their corresponding semantic regions of the label $L^x$.
We experimentally demonstrate that our style harmonization layer adaptively controls the style of each object well, and the results are given in the experiments section.

\subsection{Loss Functions} \label{3_loss_functions}
We leverage standard loss functions used in MUNIT~\cite{huang2018multimodal} to generate proper target domain images. It includes self-reconstruction $L_{self}$~\cite{zhu2017unpaired}, cycle consistency $L_{cycle}$~\cite{zhu2017unpaired}, perceptual $L_{perc}$~\cite{johnson2016perceptual} and adversarial loss $L_{adv}$~\cite{goodfellow2014generative}.
The detailed explanations of those loss functions are given in the supplementary material.

In this paper, we propose two advanced loss functions to facilitate style harmonization: content contrastive loss and style contrastive loss.
It is used with the aforementioned standard loss functions jointly.
In what follows, we introduce the proposed two loss functions.

\begin{figure*}[h]
	\centering
	\includegraphics[width=\linewidth]{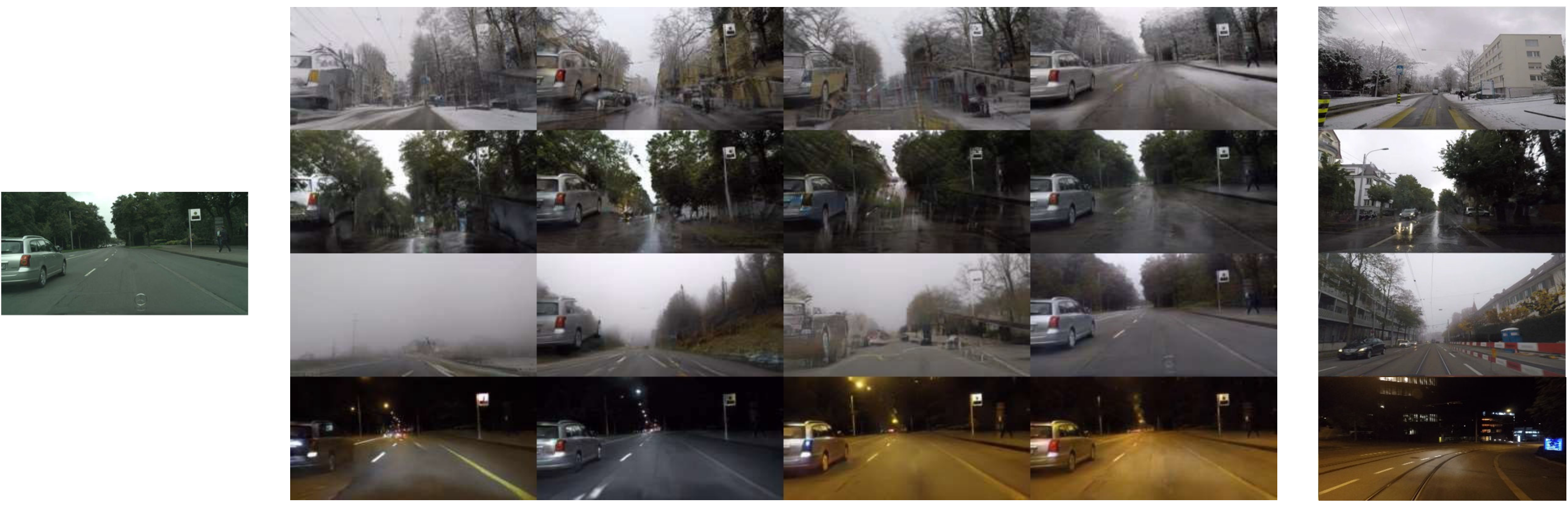}\\
	\raggedright{\scriptsize\hspace{33pt} Input \hspace{70pt} CycleGAN \hspace{45pt} MUNIT \hspace{52pt} MGUIT \hspace{42pt} SHUNIT (ours) \hspace{45pt} Target domain}
	\caption{\textbf{Qualitative comparison on Cityscapes (clear) $\rightarrow$ ACDC (snow/rain/fog/night).} From the given clear image (first column), we generate four adverse condition images using \cite{zhu2017unpaired,huang2018multimodal,jeong2021memory}
	and SHUNIT. In the last column, we show a sample of the real image for each adverse condition.}
	\label{fig:c2a}
\end{figure*}

\begin{table*}[h]
\centering
{
\footnotesize
\begin{tabular}{l|cc|cc|cc|cc}
\toprule
         & \multicolumn{2}{c|}{clear $\rightarrow$ snow} & \multicolumn{2}{c|}{clear $\rightarrow$ rain} & \multicolumn{2}{c|}{clear $\rightarrow$ fog} & \multicolumn{2}{c}{clear $\rightarrow$ night} \\
         \cmidrule{2-9} 
         & cFID $\downarrow$           & mIoU $\uparrow$           & cFID $\downarrow$           & mIoU $\uparrow$           & cFID $\downarrow$           & mIoU $\uparrow$          & cFID $\downarrow$           & mIoU $\uparrow$           \\ 
\cmidrule{1-9}\morecmidrules \cmidrule{1-9}
CycleGAN~\cite{zhu2017unpaired} & 21.88          & 23.68          & 20.16          & 35.96          & 31.72          & 13.73         & 15.26          & 31.33          \\
UNIT~\cite{liu2017unsupervised}     & 13.89          & 31.24          & 16.25          & 39.39          & 29.36          & 28.70         & \underline{12.28}          & 35.29          \\
MUNIT~\cite{huang2018multimodal}    & 13.79          & 33.83          & 12.62           & 44.20          & 29.34          & 27.44         & 12.56          & \textbf{37.43}          \\
TSIT~\cite{jiang2020tsit}     & 10.47           & \underline{38.08}          & 14.16           & \underline{46.40}          & 25.16          & \underline{36.68}         & \textbf{11.62}          & \underline{35.92}          \\
MGUIT~\cite{jeong2021memory}    & \underline{8.75}           & 33.33          & \underline{10.76}           & 42.60          & \underline{24.36}          & 10.22         & 15.83          & 31.36          \\
SHUNIT (ours)     & \textbf{6.62}           & \textbf{45.15}          & \textbf{8.47}           & \textbf{48.84}          & \textbf{6.53}           & \textbf{38.96}         & 14.08          & 33.66          \\ \hline
\end{tabular}
\caption{\label{table:c2a}\textbf{Quantitative comparison on Cityscapes $\rightarrow$ ACDC.} We measure class-wise FID (lower is better) and mIoU (higher is better). For brevity, class-wise FID is written as cFID.}
}
\end{table*}

\begin{table*}[hbt!]
\centering
{
\footnotesize
\begin{tabular}{l|c|c|c|c}
\toprule
         & clear $\rightarrow$ snow           & clear $\rightarrow$ rain           & clear $\rightarrow$ fog           & clear $\rightarrow$ night\\
\cmidrule{1-5}\morecmidrules \cmidrule{1-5}
AdaptSegNet~\cite{tsai2018learning} & 35.3          & 49.0         & 31.8          & 29.7            \\
ADVENT~\cite{vu2019advent}     & 32.1          & 44.3          & 32.9          & 31.7          \\
BDL~\cite{li2019bidirectional}    & 36.4          & \underline{49.7}          & 37.7           & \underline{33.8}                  \\
CLAN~\cite{luo2019taking}     & 37.7           & 44.0          & \underline{39.0}           & 31.6           \\
FDA~\cite{yang2020fda}     & \textbf{46.9}           & \textbf{53.3}          & \textbf{39.5}           & \textbf{37.1}            \\
SIM~\cite{wang2020differential}     & 33.3           & 44.5          & 36.6           & 28.0            \\
MRNet~\cite{zheng2021rectifying}     & 38.7           & 45.4          & 38.8           & 27.9        \\
SHUNIT (ours)     & \underline{45.2}          & 48.8          & \underline{39.0}           & 33.7          \\ \bottomrule
\end{tabular}
}
\caption{\label{table:segmentation_DA}\textbf{Quantitative Comparison on domain adaptation for semantic segmentation.} We report mIoU for Cityscapes $\rightarrow$ ACDC.}
\end{table*}

\paragraph{Content contrastive loss.}
To extract domain invariant content features from the content encoder, MUNIT~\cite{huang2018multimodal} simply reduces the L1 distances between $c^x$ and $\hat{c}^y$.
We replace this with contrastive representation learning to improve discrimination within a class.
For a content feature $\hat{c}_{i}^y$ at pixel $i$, which is the content feature extracted from translated target image, we set the positive sample to $c_i^x$ and we set the remaining features at the other pixels as negative samples.
The content contrastive loss is defined with the a form of InfoNCE~\cite{van2018representation} as: 
\begin{equation}
    \mathcal{L}_{content} = -\sum_{i=1}^{HW} log \left(\frac{\exp((c_{i}^x \cdot \hat{c}_{i}^y) / \tau)}{\sum_{j=1}^{HW} \exp((c_{j}^x \cdot \hat{c}_{i}^y)/\tau)} \right)
\end{equation}
where $\tau$ is a temperature parameter. In this equation, the features in the same class at the pixel $i$ can be considered as negative samples. This encourages the content encoder to extract more diverse style representations from the style memory within the same class. It is also applied to the target-to-source pipeline with $\hat{c}^x$ and $c^y$

\paragraph{Style constrative loss.}
We propose the style contrastive loss to allow the style memory to learn class-wise style representations.
Similar to the content contrastive loss, for a style feature $\hat{s}_{i}^x$ at pixel $i$, which is the memory style of target-to-source mapping, we set the positive sample to the source component style $s_{i}^x$ and we set the remaining features at the other pixels as negative samples.
The style constrative loss is defined as follows: 
\begin{equation}
    \mathcal{L}_{style} = -\sum_{i=1}^{HW} log \left(\frac{\exp((s_{i}^x \cdot \hat{s}_{i}^x)/\tau)}{\sum_{j=1}^{HW} \exp((s_{j}^x \cdot \hat{s}_{i}^x)/\tau)} \right)
\end{equation}
This loss directly supervises the memory style $\hat{s}^x$ for the translated target image.
This improves the stability of cycle consistency learning for reconstructing $I^x$ from $\hat{I}^y$. It is also applied to the target styles, \textit{i.e.}, $\hat{s}^y$ and $s^y$

Finally, all loss functions are summarized as follows:
\begin{equation}\label{eq:loss}
    \begin{aligned}
        \underset{(E^x, E^y, G^x, G^y)}{min} \underset{(D^x, D^y)}{max} \mathcal{L}(E^x, E^y, G^x, G^y, D^x, D^y) = \\ \lambda_{self}\mathcal{L}_{self} + \lambda_{cycle}\mathcal{L}_{cycle}+ \lambda_{perc}\mathcal{L}_{perc} \\ + \lambda_{adv}\mathcal{L}_{adv} +  \lambda_{content}\mathcal{L}_{content}+ \lambda_{style}\mathcal{L}_{style}
    \end{aligned}
\end{equation}
where $D^x$ and $D^y$ denote the multi-scale discriminators~\cite{wang2018high} for each visual domain, $\mathcal{X}$ and $\mathcal{Y}$. 
The details of $\mathcal{L}_{self}$, $\mathcal{L}_{cycle}$, $\mathcal{L}_{perc}$, and $\mathcal{L}_{adv}$ are described in the supplementary material.

\section{Experiments} \label{sec:experiments}
In this section, we present extensive experimental results and analysis.
To demonstrate the superiority of our method, we compare our SHUNIT with state-of-the-art I2I translation methods.
The implementation details of our method are provided in the supplementary material.

\begin{figure*}[t]
	\centering
	\includegraphics[width=\linewidth]{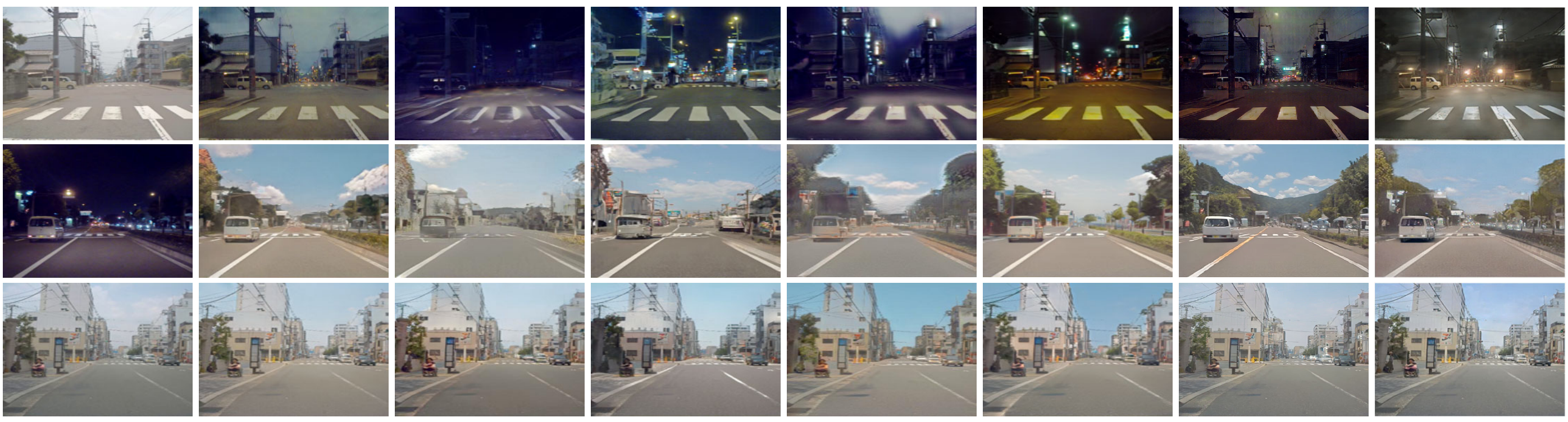}\\
	\raggedright{\scriptsize\hspace{22pt} Input \hspace{37pt} CycleGAN \hspace{37pt} UNIT \hspace{40pt} MUNIT \hspace{40pt} DRIT \hspace{40pt} MGUIT \hspace{32pt} InstaFormer \hspace{20pt} SHUNIT (ours)}
	\caption{\textbf{Qualitative comparison on INIT dataset.} (Top to bottom) sunny$\rightarrow$night, night$\rightarrow$sunny, cloudy$\rightarrow$sunny results. Our method preserves object details and looks more realistic.}
	\label{fig:init}
\end{figure*}

\begin{table*}[h]
\centering
\resizebox{\textwidth}{!}
{
\footnotesize
\begin{tabular}{l|cc|cc|cc|cc|cc|cc}
\toprule
         & \multicolumn{2}{c|}{sunny $\rightarrow$ night} & \multicolumn{2}{c|}{night $\rightarrow$ sunny} & \multicolumn{2}{c|}{sunny $\rightarrow$ rainy} & \multicolumn{2}{c|}{sunny $\rightarrow$ cloudy} & \multicolumn{2}{c|}{cloudy $\rightarrow$ sunny} &
         \multicolumn{2}{c}{Average} \\
         \cmidrule{2-13} 
         & CIS $\uparrow$           & IS $\uparrow$           & CIS $\uparrow$           & IS $\uparrow$           & CIS $\uparrow$           & IS $\uparrow$          & CIS $\uparrow$           & IS $\uparrow$           & CIS $\uparrow$           & IS $\uparrow$   & CIS $\uparrow$   & IS $\uparrow$ \\ 
\cmidrule{1-13}\morecmidrules \cmidrule{1-13}
CycleGAN~\cite{zhu2017unpaired} & 0.014          & 1.026          & 0.012          & 1.023          & 0.011          & 1.073         & 0.014          & 1.097          &0.090 &1.033 & 0.025 & 1.057 \\
UNIT~\cite{liu2017unsupervised}     & 0.082          & 1.030          & 0.027          & 1.024          & 0.097          & 1.075         & 0.081          & 1.134          & 0.219          & 1.046 & 0.087   & 1.055\\
MUNIT~\cite{huang2018multimodal}    & 1.159          & 1.278          & 1.036           & 1.051          & 1.012          & 1.146         & 1.008          & 1.095  & 1.026   & 1.321  & 1.032  & 1.166        \\
DRIT~\cite{lee2018diverse}     & 1.058           & 1.224          & 1.024           & 1.099          & 1.007          & 1.207         & 1.025          & 1.104        & 1.046          & 1.321 & 1.031  & 1.164 \\
INIT~\cite{shen2019towards}     & 1.060           & 1.118          & 1.045           & 1.080          & 1.036          & 1.152         & 1.040          & 1.142        & 1.016          & 1.460 & 1.043   & 1.179 \\
DUNIT~\cite{bhattacharjee2020dunit}     & 1.166           & 1.259          & 1.083           & 1.108          & 1.029          & 1.225         & 1.033          & 1.149        & 1.077          & 1.472 & 1.079   &   1.223 \\
MGUIT~\cite{jeong2021memory}     & 1.176           & 1.271          & \underline{1.115}           & \underline{1.130}          & 1.092          & 1.213         & 1.052          & 1.218        & \underline{1.136}          & \underline{1.489} & 1.112 & 1.254 \\
Instaformer~\cite{kim2022instaformer}     & \underline{1.200}           & \underline{1.404}          & \underline{1.115}           & 1.127          & \textbf{1.158}          & \underline{1.394}         & \textbf{1.130}          & \underline{1.257}        & \textbf{1.141}          & \textbf{1.585} & \underline{1.149} & \underline{1.353} \\
SHUNIT (ours)     & \textbf{1.205}          & \textbf{1.503}          & \textbf{1.308}           & \textbf{1.585}          & \underline{1.136}          & \textbf{1.609}         & \underline{1.111}         & \textbf{1.405}        & 1.085          & 1.315 & \textbf{1.169}   & \textbf{1.483} \\ \bottomrule
\end{tabular}
}
\caption{\label{table:init}\textbf{Quantitative Comparison on INIT dataset.} We measure CIS and IS (higher is better).}
\end{table*}

\begin{table*}[t]
\centering
{
\footnotesize
\begin{tabular}{l|c|c|c|c|c}
\toprule
         & Pers.           & Car           & Truc.           & Bic.  & mAP\\
\cmidrule{1-6}\morecmidrules \cmidrule{1-6}
DT~\cite{inoue2018cross} & 28.5          & 40.7         & 25.9          & 29.7          & 31.2   \\
DAF~\cite{huang2018multimodal}     & 39.2          & 40.2          & 25.7          & 48.9          & 38.5  \\
DARL~\cite{kim2019diversify}    & 46.4          & 58.7          & 27.0           & 49.1          & 45.3         \\
DAOD~\cite{rodriguez2019domain}     & 47.3           & 59.1          & 28.3           & 49.6          & 46.1  \\
DUNIT~\cite{bhattacharjee2020dunit}     & \underline{60.7}           & 65.1          & 32.7           & \underline{57.7}          & 54.1  \\
MGUIT~\cite{jeong2021memory}     & 58.3           & 68.2          & 33.4           & \textbf{58.4}          & 54.6  \\
InstaFormer~\cite{kim2022instaformer}     & \textbf{61.8}           & \underline{69.5}          & \underline{35.3}           & 55.3          & \underline{55.5}  \\
SHUNIT (ours)     & 56.3          & \textbf{74.4}          & \textbf{51.9}           & 53.2          & \textbf{59.0}  \\ \bottomrule
\end{tabular}
}
\caption{\label{table:DA}\textbf{Quantitative Comparison on domain adaptation for object detection.} We report per-class AP for KITTI $\rightarrow$ Cityscapes.}
\end{table*}

\subsection{Datasets}
We evaluate our SHUNIT on three I2I translation scenarios: Cityscapes \cite{cordts2016cityscapes} $\rightarrow$ ACDC \cite{sakaridis2021acdc} and INIT \cite{shen2019towards}, and KITTI~\cite{geiger2013vision} $\rightarrow$ Cityscapes~\cite{cordts2016cityscapes}.
In all scenarios, INIT~\cite{shen2019towards}, DUNIT~\cite{bhattacharjee2020dunit}, MGUIT~\cite{jeong2021memory}, InstaFormer~\cite{kim2022instaformer}, and our method use semantic labels provided in each dataset.

\subsubsection{Cityscapes $\rightarrow$ ACDC} Cityscapes \cite{cordts2016cityscapes} is one of the most popular urban scene dataset. ACDC \cite{sakaridis2021acdc} is the latest dataset with multiple adverse condition images and consists of four conditions of street scenes: snow, rain, fog, and night. ACDC dataset provides images with corresponding dense pixel-level semantic annotations, and it has 19 classes the same as Cityscapes dataset for all adverse conditions. Following \cite{sakaridis2021acdc}, we leverage Cityscapes dataset as a clear condition and translate it to the adverse conditions (\textit{i.e.}, snow, rain, fog, and night) in ACDC dataset. Therefore, this scenario is challenging because not only the weather conditions, but also layouts, such as camera model, view, and angle, are different. To train the networks, 2975, 400, 400, 400, and 400 images are used for clear, snow, rain, fog, and night conditions, respectively. For a fair comparison on this benchmark, we reproduce existing state-of-the-art methods \cite{zhu2017unpaired, liu2017unsupervised, huang2018multimodal, jiang2020tsit, jeong2021memory} in our system. For fair comparison, we set the number of key-value pairs for style memory to be the same as our setting and use segmentation mask for reproducing ~\cite{jeong2021memory}.

\subsubsection{INIT} INIT \cite{shen2019towards} is a public benchmark set for I2I translation. It contains street scenes images including 4 weather categories (\textit{i.e.}, sunny, night, rainy, and cloudy) with the corresponding bounding box labels.
Following \cite{shen2019towards}, we split the 155K images into 85\% for training and 15\% for testing. We conduct five translation experiments: sunny $\leftrightarrow$ night, sunny $\leftrightarrow$ cloudy, sunny $\rightarrow$ rainy. In this dataset, we directly copied the results of the existing methods from \cite{shen2019towards, bhattacharjee2020dunit, jeong2021memory, kim2022instaformer}. Similarly, for fair comparison with MGUIT, the number of key-value pairs in style memory is set equally.

\subsubsection{KITTI $\rightarrow$ Cityscapes}
KITTI is a public benchmark set for object detection. It contains 7481 images with bounding boxes annotations for training and 7518 images for testing. Following the previous I2I translation methods~\cite{bhattacharjee2020dunit, jeong2021memory, kim2022instaformer}, we select the common 4 object classes (person, car, truck, bicycle) for evaluatation.

\begin{table*}
  \setlength\tabcolsep{4pt}
  \centering
  \hfill%
  \subfloat[\label{table:ablation1}Style ablation study on SHL.]{%
    \resizebox{!}{1.3cm}{
    \begin{tabular}{ccc|cc|cc}
    \toprule
               &            &            & \multicolumn{2}{c|}{clear $\rightarrow$ snow} & \multicolumn{2}{c}{clear $\rightarrow$ rain} \\ \cmidrule{4-7} 
    Mem.       & Comp.       & $\alpha$   & cFID $\downarrow$      & mIoU $\uparrow$      & cFID $\downarrow$      & mIoU $\uparrow$     \\ \cmidrule{1-7} \morecmidrules \cmidrule{1-7}
    \checkmark &            &            & 12.93                   & 26.33                & \textbf{7.83}          & 39.06               \\
    \checkmark & \checkmark &            & 8.72                   & 37.20                 & 8.47                   & 38.97               \\
    \checkmark & \checkmark & \checkmark & \textbf{6.62}          & \textbf{44.35}       & 8.47                   & \textbf{42.77}     \\
    \bottomrule
    \end{tabular}}}%
  \hfill%
  \subfloat[\label{table:ablation3}Ablation study on loss functions.]{%
    \resizebox{!}{1.3cm}{
    \begin{tabular}{cc|cc|cc}
    \toprule
                            &                       & \multicolumn{2}{c|}{clear $\rightarrow$ snow} & \multicolumn{2}{c}{clear $\rightarrow$ rain} \\ \cmidrule{3-6} 
    $\mathcal{L}_{content}$ & $\mathcal{L}_{style}$ & cFID $\downarrow$      & mIoU $\uparrow$      & cFID $\downarrow$      & mIoU $\uparrow$     \\ \cmidrule{1-6} \morecmidrules \cmidrule{1-6}
    \checkmark              &  & 11.06           & 32.43          & 13.23          & 38.39          \\
                  & \checkmark   & 12.38           & 30.89          & 9.33          & 39.47          \\ 
    \checkmark              & \checkmark            & \textbf{6.62}           & \textbf{44.35}          & \textbf{8.47}          & \textbf{42.77}          \\
    \bottomrule
    \end{tabular}}}%
  \hspace{\fill}

  \hfill%
  \subfloat[\label{table:ablation2}Style memory training strategies.]{%
    \resizebox{!}{1.0cm}{
    \begin{tabular}{l|cc|cc}
    \toprule
                       & \multicolumn{2}{c|}{clear $\rightarrow$ snow} & \multicolumn{2}{c}{clear $\rightarrow$ rain} \\ \cmidrule{2-5} 
                       & cFID $\downarrow$           & mIoU $\uparrow$           & cFID $\downarrow$          & mIoU $\uparrow$           \\ \cmidrule{1-5} \morecmidrules \cmidrule{1-5}
    Updating & 10.55           & 36.70          & 12.05          & 38.21          \\
    Backprop.             & \textbf{6.62}  & \textbf{44.35} & \textbf{8.47} & \textbf{42.77} \\ 
    \bottomrule
    \end{tabular}}}%
  \hfill%
  \subfloat[\label{table:ablation4}L1 \textit{vs.} Contrastive loss for style and content losses.]{%
    \resizebox{!}{1.0cm}{
    \begin{tabular}{l|cc|cc}
    \toprule
                        & \multicolumn{2}{c|}{clear $\rightarrow$ snow} & \multicolumn{2}{c}{clear $\rightarrow$ rain} \\ \cmidrule{2-5} 
                        & cFID $\downarrow$           & mIoU $\uparrow$           & cFID $\downarrow$          & mIoU $\uparrow$           \\ \cmidrule{1-5} \morecmidrules \cmidrule{1-5}
    L1 & 12.21           & 38.98          & \textbf{7.72} & 39.78          \\ 
    Contrastive              & \textbf{6.62}  & \textbf{44.35} & 8.47          & \textbf{42.77} \\ 
    \bottomrule
    \end{tabular}}}%
  \hfill%
  \subfloat[\label{table:ablation5}Experimental results of label input for content encoder.]{%
    \resizebox{!}{1.0cm}{
    \begin{tabular}{l|cc|cc}
    \toprule
                        & \multicolumn{2}{c|}{clear $\rightarrow$ snow} & \multicolumn{2}{c}{clear $\rightarrow$ rain} \\ \cmidrule{2-5} 
                           & cFID $\downarrow$           & mIoU $\uparrow$           & cFID $\downarrow$          & mIoU $\uparrow$           \\ \cmidrule{1-5} \morecmidrules \cmidrule{1-5}
    w/o & 6.73           & 44.11          & 9.03           & 38.72          \\
    w/                  & \textbf{6.62}           & \textbf{44.35}          & \textbf{8.47}          & \textbf{42.77}          \\
    \bottomrule
    \end{tabular}}}%
  \hspace{\fill}
\caption{\textbf{Ablation Study.} We report class-wise FID and mIoU in two scenarios: clear $\rightarrow$ \{snow, rain\}.}
\end{table*}

\subsection{Qualitative Comparison} \label{4_qual}
Fig.~\ref{fig:c2a} shows qualitative results on Cityscapes $\rightarrow$ ACDC.
Since our I2I translation setting, Cityscapes $\rightarrow$ ACDC, is very challenging as discussed in the datasets section, existing methods cannot generate realistic images in several scenarios. Specifically, CycleGAN~\cite{zhu2017unpaired} often destroys the semantic layout. MUNIT~\cite{huang2018multimodal} translates images with a global style, thus it also often generates artifacts, as shown in the snow, rain, and fog images. MGUIT~\cite{jeong2021memory} also includes artifacts in the car even though leveraging memory style. It shows the limitation of the updating mechanism for training memory style, and the limitation is clearly depicted in the challenging scenario. In contrast to them, our SHUNIT accurately generates images in the target domains without losing the original style in the input image. In the supplementary material, we further provide the results of UNIT~\cite{liu2017unsupervised} and TSIT~\cite{jiang2020tsit}.

As shown in Fig.~\ref{fig:init}, which depicts qualitative results on INIT dataset, our method generates high-quality images in various scenarios. In the night $\rightarrow$ sunny scenario (second row), InstaFormer~\cite{kim2022instaformer} translates the color of the road lane to yellow. On the other hand, our method keeps the color of the lane as white and generates a sunny scene by harmonizing the target domain style retrieved from a style memory and an image style.

\subsection{Quantitative Comparison} \label{4_quan}
The quantitative results on Cityscapes $\rightarrow$ ACDC are presented in Table~\ref{table:c2a}. 
To quantify the per-class image-to-image translation quality, we measure class-wise FID~\cite{shim2022local}.
We further measure mIoU on ACDC test set.
The mIoU metric is used to validate the results on the practical problem, semantic segmentation.
We generate a training set by Cityscapes $\rightarrow$ ACDC and then train DeepLabV2~\cite{chen2017deeplab} on it.
The mIoU score is obtained with the trained DeepLabV2 by evaluating on ACDC test set.
As shown in Table~\ref{table:c2a}, we surpass the state-of-the-art I2I translation methods by a significant margin in most scenarios, demonstrating the superiority of our style harmonization for unpaired I2I translation.
Table~\ref{table:segmentation_DA} shows the quantitative results of domain adaptation for semantic segmentation with the state-of-the-art methods. Despite we trained DeepLabV2 with only a simple cross-entropy loss, our SHUNIT achieves comparable performance with the domain adaptation methods that were trained DeepLabV2 with additional loss functions and several techniques for boosting the performance of mIoU score on the target domain.

Table~\ref{table:init} shows another quantitative results on the testing split of INIT \cite{shen2019towards}.
To directly compare our method with the public results, we evaluate our method with Inception Score (IS) \cite{salimans2016improved} and Conditional Inception Score (CIS) \cite{huang2018multimodal}.
As shown in Table~\ref{table:init}, we achieve the best performance in most scenarios.

We further evaluate our method on domain adaptation benchrmark following DUNIT~\cite{bhattacharjee2020dunit}. We use Faster-RCNN~\cite{ren2015faster} trained on the source domain as a detector. As shown in Table~\ref{table:DA}, we achieve the state-of-the-art performance.

\subsection{Ablation Study} \label{4_ablation}
In this section, we study the effectiveness of each component in our method.
We validate on Cityscapes (clear) $\rightarrow$ ACDC (snow/rain) scenarios and use ACDC validation set for both class-wise FID and mIoU\footnote{mIoU on test set should be evaluated on the online server \cite{sakaridis2021acdc} and it has a limit on the number of submissions. Therefore, we use validation set for ablation study.}.

\paragraph{Ablation study on style harmonization layer.}
We ablate the memory style, component style, and class-wise $\alpha$ in the style harmonization layer, and they are denoted as ``Mem.'', ``Comp.'', and ``$\alpha$'' in Table~\ref{table:ablation1}, respectively.
As shown in the table, the memory style-only is far behind the full model. With component style, we can achieve performance improvement on clear $\rightarrow$ snow while decreasing on clear $\rightarrow$ rain. We obtain significant improvement on most scenarios with class-wise $\alpha$. The results demonstrate that the existing approach, which only leverages the memory style, is not sufficient for I2I translation, and we successfully address the problem by adaptively harmonizing two styles.

\paragraph{Ablation study on content and style losses.}
We study the effectiveness of the proposed two loss functions, $\mathcal{L}_{style}$ and $\mathcal{L}_{content}$, by ablating them step-by-step, and the results are given in Table~\ref{table:ablation3}. As shown in the table, our model is effective when two losses are used jointly.

\paragraph{Style memory training strategy.}
As described in the proposed method section, we opt for backpropagation to train the style memory rather than updating the mechanism used in \cite{jeong2021memory}.
The results are shown in Table~\ref{table:ablation2}.
We surpass the existing updating method by a large margin.

\paragraph{L1 \textit{vs.} Contrastive loss.}
Table~\ref{table:ablation4} shows the efficacy of our contrastive-based approach by replacing $L_{content}$ and $L_{style}$ with L1 losses as used in MUNIT~\cite{huang2018multimodal}.
The L1 losses are designed to reduce L1 distances within positive pairs without consideration of negative pairs.
As we discussed in loss functions section, our method effectively encourages extracting more diverse style representations, leading to performance improvement.

\paragraph{Label input for content encoder.}
Table~\ref{table:ablation5} shows that label input is not significant but always leads to performance improvements.
Therefore, we have no reason to omit the label input.

\subsection{Limitations}
Since our framework leverages the component style of the source image, the generated image's quality relies on the source image's quality.
If the source image has a very bright colors, SHUNIT often generates a relatively bright night image. In Fig.~\ref{fig:c2a}, SHUNIT struggles to generate geometrically distinct lights from the source image. Due to the above problems, SHUNIT cannot achieve the best performance on clear $\rightarrow$ night scenario in Table.~\ref{table:c2a}. We believe that these problems can be alleviated by leveraging geometric information such as depth or camera pose.
Additionally, our approach can give limited benefit to some I2I translation scenarios, such as dog $\rightarrow$ cat, because these tasks need to change the content; however, we tackle the unpaired I2I translation task under the condition that the content will not be changed, and only the style will be changed.

\section{Conclusion} \label{sec:conclusion}
We present a new perspective of the target style: It can be disentangled into class-aware and image-specific styles. Furthermore, our SHUNIT effectively harmonizes the two styles, and its superiority is demonstrated through extensive experiments. We believe that our proposal has the potential to break new ground in style-based image editing applications such as style transfer, colorization, and image inpainting.

\section{Acknowledgement}
This work was supported by Institute of Information \& communications Technology Planning \& Evaluation (IITP) grant funded by the Korea government (MSIT) (No.2021-0-00800, Development of Driving Environment Data Transformation and Data Verification Technology for the Mutual Utilization of Self-driving Learning Data for Different Vehicles).

\bibliography{main}

\clearpage

\appendix


\section{Implementation Details} \label{sec:1}

\subsection{Experiments Settings}
We implement our model with 1.7.1 version of  PyTorch~\cite{paszke2017automatic} framework.
To train SHUNIT, we use a fixed learning rate of $10^{-4}$ and use the Adam optimizer~\cite{kingma2014adam} with $\beta_1$ and $\beta_2$ of 0.5 and 0.999, respectively.
The weight decay is set to $10^{-4}$.
Our models are trained with a single NVIDIA RTX A6000 GPU.
For the experiments on Cityscapes~\cite{cordts2016cityscapes} $\rightarrow$ ACDC~\cite{sakaridis2021acdc}, we use RGB images with a size of $512 \times 256$ during both training and inference. 
We train the model for 50K iterations.
Since the ACDC dataset has been published recently, no official results are available on this dataset for the existing methods.
Therefore, we reproduced CycleGAN~\cite{zhu2017unpaired}, UNIT~\cite{liu2017unsupervised}, MUNIT~\cite{huang2018multimodal}, TSIT~\cite{jiang2020tsit}, and MGUIT~\cite{jeong2021memory} with their official implementation code\textsuperscript{\rm 1, 2, 3, 4}.
\renewcommand{\thefootnote}{}
\footnote{\textsuperscript{\rm 1}The code of CycleGAN was taken from \\ https://github.com/junyanz/pytorch-CycleGAN-and-pix2pix}
\footnote{\textsuperscript{\rm 2}The code of UNIT and MUNIT were taken from \\ https://github.com/NVlabs/MUNIT}
\footnote{\textsuperscript{\rm 3}The code of TSIT was taken from \\ https://github.com/EndlessSora/TSIT}
\footnote{\textsuperscript{\rm 4}The code of MGUIT was taken from \\ https://github.com/somijeong/MGUIT}

For the experiments on INIT~\cite{shen2019towards}, we use RGB images with a size of $360 \times 360$ during training while we use RGB images with a size of $572 \times 360$ during inference.
We train the model for 250K iterations.
For the experiment on KITTI~\cite{geiger2013vision} $\rightarrow$ Cityscapes, we use RGB images with a size of $540 \times 360$ during training while we use RGB images with a size of $1242 \times 375$ during inference. We train the model for 200K iterations.

For fair comparisons with INIT~\cite{shen2019towards}, DUNIT~\cite{bhattacharjee2020dunit}, MGUIT~\cite{jeong2021memory}, and InstaFormer~\cite{kim2022instaformer}, which used bounding-box labels as an additional object annotation, we did not use segmentation labels but used bounding-box labels for all experiments on both INIT and KITTI $\rightarrow$ Cityscapes scenarios. In addition, we reproduced MGUIT~\cite{jeong2021memory} on Cityscapes $\rightarrow$ ACDC experiments with segmentation labels.

\subsection{Details of Network Architecture}
In the content encoder $E_c^x$, we employ two convolutional networks, one network takes RGB input, and the other takes one-hot encoded semantic label input.
Each network consists of Conv(input channel, 64, 7, 1, 3, IN, relu) - Conv(64, 128, 4, 2, 1, IN, relu) - Conv(128, 256, 4, 2, 1, IN, relu) - Resblk~$\times$4 - Conv(256, 128, 1, 1, IN, relu), where IN denotes the instance normalization layer~\cite{ulyanov2016instance}; Resblk denotes the residual block~\cite{he2016deep}; and the components in Conv($\cdot$) denotes (input channel, output channel, kernel size, stride, padding, normalization, activation).
The encoded RGB and semantic label features are then concatenated along the channel dimension, and it is the content feature $c^x$.
The style encoder $E_s^x$ has the same network architecture as a convolutional network in the content encoder, except for the normalization layer.
The style encoder does not use normalization layers to keep the original style of the input image.
The generator $G^y$ consists of SH~Resblk~$\times$4 - Conv(256, 128, 5, 1, 2, IN, relu) - Conv(128, 64, 5, 1, 2, IN, relu) - Conv(64, 3, 7, 1, 3, -, tanh).
For the experiments on Cityscapes $\rightarrow$ ACDC, we use 20 key-value pairs for each class in the style memory $M_n^y$. 
In the rest of the experiments, we follow key-value pairs of MGUIT~\cite{jeong2021memory}.

\subsection{Details of Loss Functions}
Following MUNIT~\cite{huang2018multimodal}, we leverage self-reconstruction $\mathcal{L}_{self}$~\cite{zhu2017unpaired}, cycle consistency $\mathcal{L}_{cycle}$~\cite{zhu2017unpaired}, perceptual $\mathcal{L}_{perc}$~\cite{johnson2016perceptual}, and adversarial $\mathcal{L}_{adv}$~\cite{goodfellow2014generative} losses as follows:

\begin{equation}
    \mathcal{L}_{self} = \left\|I^x - G^x(c^x, s^x)\right\|_{1} + \left\|I^y - G^y(c^y, s^y)\right\|_{1},
\end{equation}

\begin{equation}
    \mathcal{L}_{cycle} = \left\|I^x - G^x(\hat{c}^y, \hat{s}^x)\right\|_{1} + \left\|I^y - G^y(\hat{c}^x, \hat{s}^y)\right\|_{1},
\end{equation}

\begin{equation}
    \mathcal{L}_{perc} = \left\|F(I^x) - F(\hat{I^y})\right\|_{1} + \left\|F(I^y) - F(\hat{I^x})\right\|_{1},
\end{equation}

\begin{equation}
\begin{split}
    \mathcal{L}_{adv} = \left [\log{(1-D^x(\hat{I}^x))} + \log{D^x(I^x)}\right ] \\  + \left [\log{(1-D^y(\hat{I}^y))} + \log{D^y(I^y)}\right ],
\end{split}
\end{equation}
where $F(\cdot)$ is a feature map extracted from \texttt{relu5\_3} layer of the ImageNet pretrained VGG-16 network~\cite{simonyan2014very}; $D^{x}(\cdot)$ and $D^{y}(\cdot)$ are features extracted from the multi-scale discriminators~\cite{wang2018high} for each visual domain, $\mathcal{X}$ and $\mathcal{Y}$, respectively.
We empirically set the hyperparameters \{$\lambda_{self}, \lambda_{cycle}, \lambda_{perc}, \lambda_{adv}, \lambda_{content}, \lambda_{style}$\} to \{$10, 10, 1, 1, 10, 10$\}.
The temperature $\tau$ in $\mathcal{L}_{content}$ and $\mathcal{L}_{style}$ is set to $0.7$.

\section{Analysis of class-wise FID} \label{sec:2}
Recent work~\cite{shim2022local} introduced class-wise FID (cFID) to quantify the per-class image quality in the image generation task. We realized that the I2I translation task also has advantages if we evaluate the results with cFID.
In this section, we provide a detailed analysis of the necessity for cFID on our benchmark set and the implementation detail of cFID for I2I translation.

\begin{figure*}[t]
	\centering
	\includegraphics[width=0.5\linewidth]{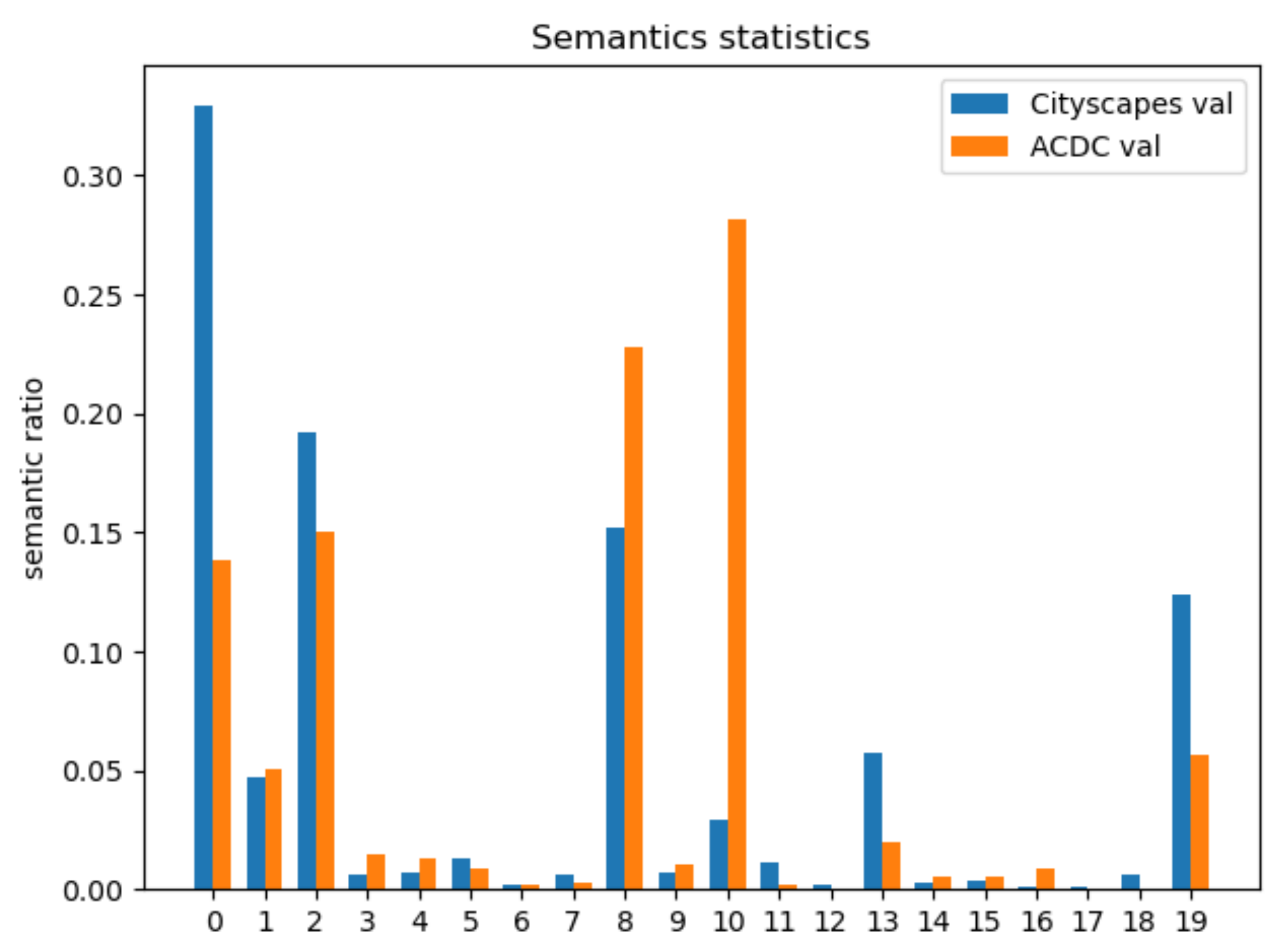}
	\caption{\textbf{The class statistics in Cityscapes~\cite{cordts2016cityscapes} validation set \textit{vs}. ACDC~\cite{sakaridis2021acdc} validation set.} Even though the two datasets share the class categories, the class statistics are naturally different.}
	\label{fig:semantic_sta}
\end{figure*}

\paragraph{Necessity of CFID.}
By the nature of unsupervised I2I, as shown in Fig.~\ref{fig:semantic_sta}, the class statistics between the source domain (Cityscapes~\cite{cordts2016cityscapes}) and the target domain (ACDC~\cite{sakaridis2021acdc}) are unmatched.
However, FID ignores the unmatched class statistics because FID only uses the global embeddings to compute the distance.
Therefore, the different class statistics prevent FID improvement, regardless of the generated image quality.

\begin{figure*}[t]
	\centering
	\includegraphics[width=\linewidth]{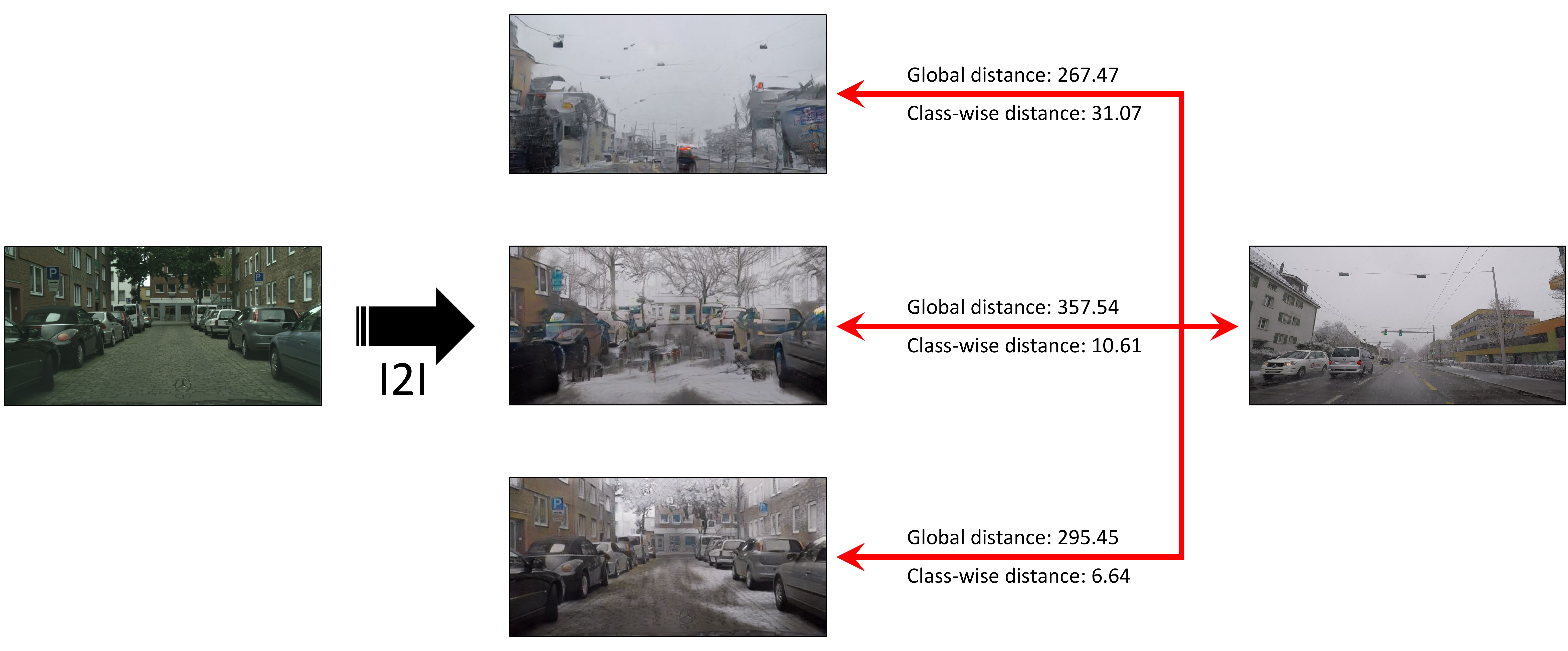}\\
	\vspace{-5.4cm}
	\raggedright{\small\hspace{155pt} (a) CycleGAN~\cite{zhu2017unpaired}}\\
	\vspace{2.18cm}
	\raggedright{\small\hspace{42pt} Input \hspace{90pt} (b) MGUIT~\cite{jeong2021memory} \hspace{148pt} (d) Target domain}\\
	\vspace{2.18cm}
	\raggedright{\small\hspace{180pt} (c) SHUNIT (ours)}\\
	\caption{\textbf{Illustration of comparison of global distance and class-wise distance.} From a clear image, we generate snow images using (a) CycleGAN~\cite{zhu2017unpaired}, (b) MGUIT~\cite{jeong2021memory}, and (c) SHUNIT. To quantitatively evaluate the results, we compute distances between (a,b,c) the generated images and (d) real snow image in two methods: One is computed with global embeddings of Inception-V3 and the other one is computed with class-wise embeddings.}
	\label{fig:cfid_ex}
\end{figure*}

To demonstrate the problem of FID intuitively, we illustrate an example in Fig.~\ref{fig:cfid_ex}.
In the figure, we translate a clear image sampled from Cityscapes dataset to snow with three methods: (a) CycleGAN~\cite{zhu2017unpaired}, (b) MGUIT~\cite{jeong2021memory}, and (c) SHUNIT.
As can be compared qualitatively, SHUNIT generates a more realistic snow image than CycleGAN.
For a quantitatively comparison, we take a similar strategy used in FID: We extract global embeddings from a real snow image sampled from ACDC dataset and the generated image using Inception-V3~\cite{szegedy2016rethinking}. Then, we compute the squared Euclidean distance between the two global embeddings.
However, as depicted in Fig.~\ref{fig:cfid_ex}, CycleGAN (267.47) achieves better performance than SHUNIT (295.45) in global distance.
The reason is that CycleGAN destroys the layout in the input image and follows the layout of the real snow image, resulting in severe artifacts in the generated image while better performance in the global distance.
If we extract class-wise embeddings and compute class-wise distance then average them, we can obtain a reasonable rank: SHUNIT (6.64) achieves the best, MGUIT (10.61) achieves the second-best, and CycleGAN (31.07) achieves the third.
As can be observed in this example, comparing class-wise embeddings is a more reasonable method to evaluate the generated quality than comparing global embeddings in unsupervised I2I translation.

\begin{figure*}[t]
	\centering
	\includegraphics[width=\linewidth]{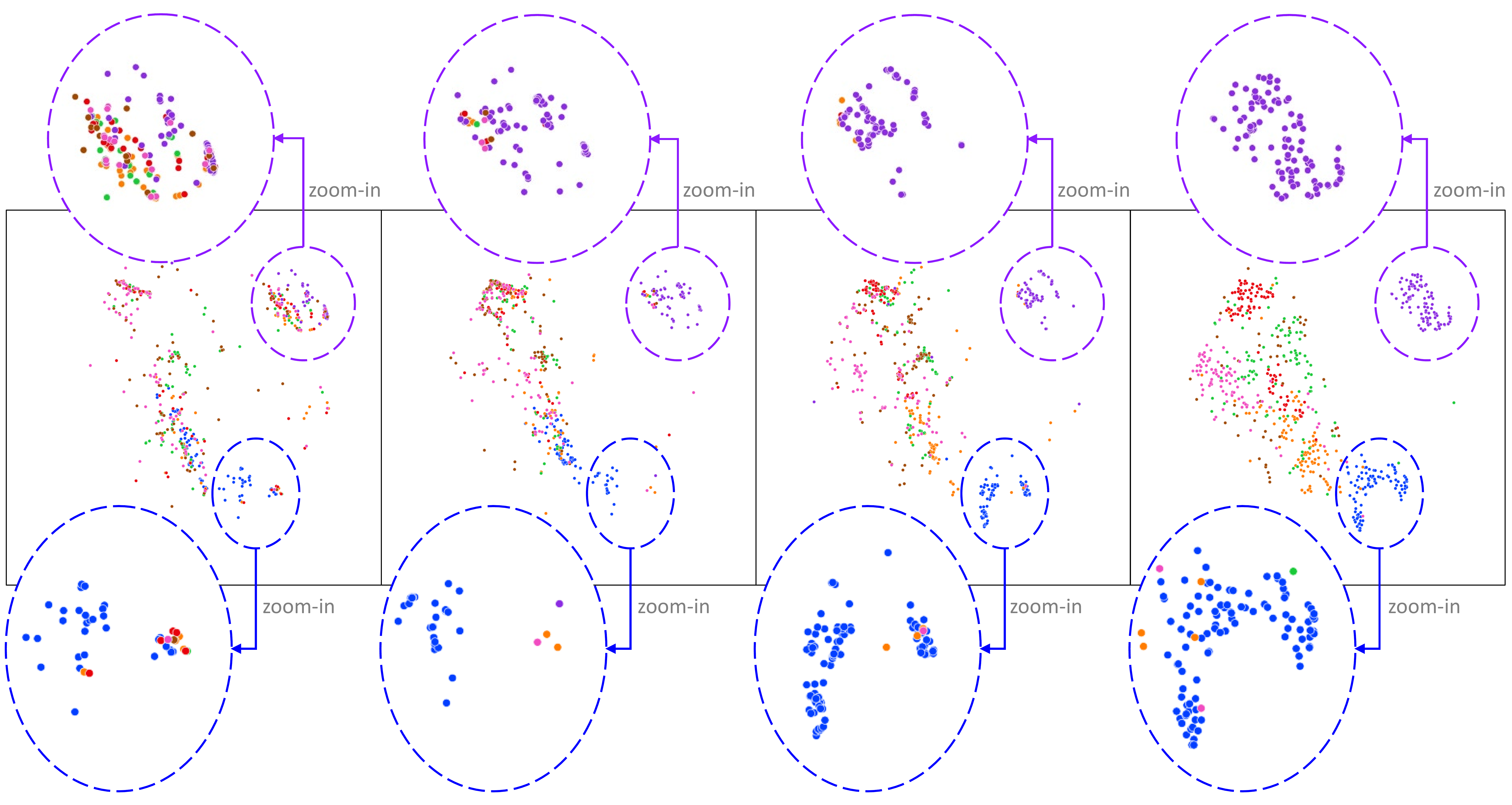}\\
	\raggedright{\hspace{1pt} (a) CycleGAN~\cite{zhu2017unpaired} \hspace{5pt} (b) MGUIT~\cite{jeong2021memory} \hspace{13pt} (c) SHUNIT (ours) \hspace{38pt} (d) Target domain}
	\caption{\textbf{t-SNE~\cite{van2008visualizing} visualization.} We plot pixel-wise features extracted from Inception-V3~\cite{szegedy2016rethinking}. In (a), (b), and (c), the generated snow images, which are translated from clear images, are used for extracting features, while in (d), the real snow images in the ACDC validation set are used. In the figure, the same color of dots are in the same class category.}
	\label{fig:fig2}
\end{figure*}

To further demonstrate the necessity of cFID, we plot the pixel-wise features of Inception-V3~\cite{szegedy2016rethinking} in Fig.~\ref{fig:fig2}.
In Figs.~\ref{fig:fig2}(a),~\ref{fig:fig2}(b), and~\ref{fig:fig2}(c), Cityscapes validation set is used as clear images for I2I translation.
In Fig.~\ref{fig:fig2}(d), ACDC validation set is used as real snow images.
In the figure, we use the same color for the same category of the pixel-wise feature.
As shown in the zoom-in regions, pixel-wise features extracted from (d) real snow images are well separated by class categories because the pretrained network is used.
However, (a) CycleGAN includes various categories in the zoom-in region, which clearly indicates that the quality of the generated images is low.
In contrast, (c) features in SHUNIT are well separated by class categories as in (d) real snow images.
This class information should be considered to evaluate the quality of the generated images.

\begin{table*}[t]
\centering
\resizebox{1.0\textwidth}{!}
{
\setlength\tabcolsep{1pt}
\footnotesize
\begin{tabular}{l|ccc|ccc|ccc|ccc}
\toprule
              & \multicolumn{3}{c|}{clear $\rightarrow$ snow} & \multicolumn{3}{c|}{clear $\rightarrow$ rain} & \multicolumn{3}{c|}{clear $\rightarrow$ fog} & \multicolumn{3}{c}{clear $\rightarrow$ night} \\ \cmidrule{2-13}
              & FID $\downarrow$       & CFID $\downarrow$    & mIoU $\uparrow$     & FID $\downarrow$       & CFID $\downarrow$     & mIoU $\uparrow$     & FID $\downarrow$       & CFID $\downarrow$     & mIoU $\uparrow$    & FID $\downarrow$       & CFID $\downarrow$     & mIoU $\uparrow$     \\ \cmidrule{1-13}\morecmidrules \cmidrule{1-13}
CycleGAN~\cite{zhu2017unpaired}      & \textbf{134.21}    & 21.88    & 23.68    & 160.91    & 20.16    & 35.96    & \underline{154.23}    & 31.72    & 13.73   & \textbf{105.40}    & 15.26    & 31.33    \\
UNIT~\cite{liu2017unsupervised}          & \underline{136.63}    & 13.89    & 31.24    & \textbf{145.79}    & 16.25    & 39.39    & 190.47    & 29.36    & 28.70   & 116.39    & \underline{12.28}    & 35.29    \\
MUNIT~\cite{huang2018multimodal}         & 136.91    & 13.79    & 33.83    & \underline{151.48}    & 12.62    & 44.20    & 195.28    & 29.34    & 27.44   & \underline{114.94}    & 12.56    & \textbf{37.43}    \\
TSIT~\cite{jiang2020tsit}          & 162.19    & 10.47    & \underline{38.08}    & 158.12    & 14.16    & \underline{46.40}    & 197.62    & 25.16    & \underline{36.68}   & 127.09    & \textbf{11.62}    & \underline{35.92}    \\
MGUIT~\cite{jeong2021memory}         & 166.76    & \underline{8.75}     & 33.33    & 178.59    & \underline{10.76}    & 42.60    & 190.86    & \underline{24.36}    & 10.22   & 119.11    & 15.83    & 31.36    \\
SHUNIT (ours) & 143.21    & \textbf{6.62}     & \textbf{45.15}    & 158.96    & \textbf{8.47}     & \textbf{48.84}    & \textbf{153.19}    & \textbf{6.53}     & \textbf{38.96}   & 125.80    & 14.08    & 33.66    \\ \bottomrule
\end{tabular}
}
\caption{\label{table:cfid_v1}\textbf{Quantitative comparison on Cityscapes $\rightarrow$ ACDC.} We measure FID, CFID (lower is better) and mIoU (higher is better).}
\end{table*}

However, as shown in Table~\ref{table:cfid_v1}, CycleGAN achieves the best performance of FID in clear $\rightarrow$ snow scenario.
Therefore, FID score is not reliable in unsupervised I2I translation with multiple classes.
In this paper, we simply and effectively address the problem by computing class-wise distances separately.

\paragraph{Implementation detail of cFID for I2I translation.}
We measure cFID based on the official implementation of FID~\cite{Seitzer2020FID}.
cFID is calculated with the Inception-V3~\cite{szegedy2016rethinking} features.
We extract the features from the first block of the network to obtain fine-scale features, which effectively include features of small objects.
To extract class-wise embeddings, we upsample the features to their original input size by bilinear interpolation, and then semantic region pooling is applied to the features.
Here, we use the ground truth of the semantic segmentation for the semantic region pooling.
With the class-wise embeddings, we compute FID for each class separately, and then cFID is obtained by averaging them.

\section{Stability of SHUNIT}\label{sec:val}

\begin{table}[t]
\centering
\resizebox{\columnwidth}{!}
{
\footnotesize
\begin{tabular}{l|cc|cc}
\toprule
               & \multicolumn{2}{c|}{clear $\rightarrow$ snow} & \multicolumn{2}{c}{clear $\rightarrow$ rain} \\ \cmidrule{2-5} 
               & cFID $\downarrow$           & mIoU $\uparrow$           & cFID $\downarrow$           & mIoU $\uparrow$          \\ \cmidrule{1-5}\morecmidrules \cmidrule{1-5}
CycleGAN~\cite{zhu2017unpaired}       & 21.88          & 23.68          & 20.16          & 35.96         \\
UNIT~\cite{liu2017unsupervised}           & 13.89          & 31.24           & 16.25          & 39.39         \\
MUNIT~\cite{huang2018multimodal}          & 13.79          & 33.83          & 12.62          & 44.20         \\
TSIT~\cite{jiang2020tsit}           & 10.47          & 38.08          & 14.16          & 46.40         \\
MGUIT~\cite{jeong2021memory}          & 8.75           & 33.33          & 10.76          & 42.60         \\ \cmidrule{1-5}
SHUNIT (trial 1, reported in the main paper)  & {6.62}           & {45.15}          & {8.47}           & {48.84} \\
SHUNIT (trial 2) & 7.18           & 44.71           & \textcolor{red}{9.53}           & \textbf{49.14}         \\
SHUNIT (trial 3) & \textbf{6.41}         & \textbf{45.94}        & {8.44}         & \textcolor{red}{48.01}        \\
SHUNIT (trial 4) & \textcolor{red}{7.56}         & \textcolor{red}{38.41}         & \textbf{8.34}         & 48.67        \\
SHUNIT (trial 5) & 6.86           & 45.41          & 8.56           & 48.70         \\ \bottomrule
\end{tabular}
}
\caption{\textbf{Results of five trials on Cityscapes $\rightarrow$ ACDC.} We measure class-wise FID (lower is better) and mIoU (higher is better). In our result, the best results are \textbf{bold-faced} while the worst results are \textcolor{red}{red-colored}. For brevity, class-wise FID is written as cFID.\label{table:iterative}}
\end{table}

To show the stability of our model, we train SHUNIT five times with different random seeds on Cityscapes (clear) $\rightarrow$ ACDC (snow, rain), and the results are given in Table~\ref{table:iterative}.
As shown in the table, the performance gap between the best and worst results is reasonably small.
In addition, the worst result also achieves the state-of-the-art performance over previous works.
This result demonstrates that our results for quantitative comparison are not cherry-picked and actually lead to performance improvements.
Furthermore, the result promises that our strong performance can be easily reproduced.

\section{More Qualitative Results} \label{sec:3}
\subsection{Cityscapes $\rightarrow$ ACDC}
\begin{figure*}[t]
	\hfill
	\includegraphics[width=.87\linewidth]{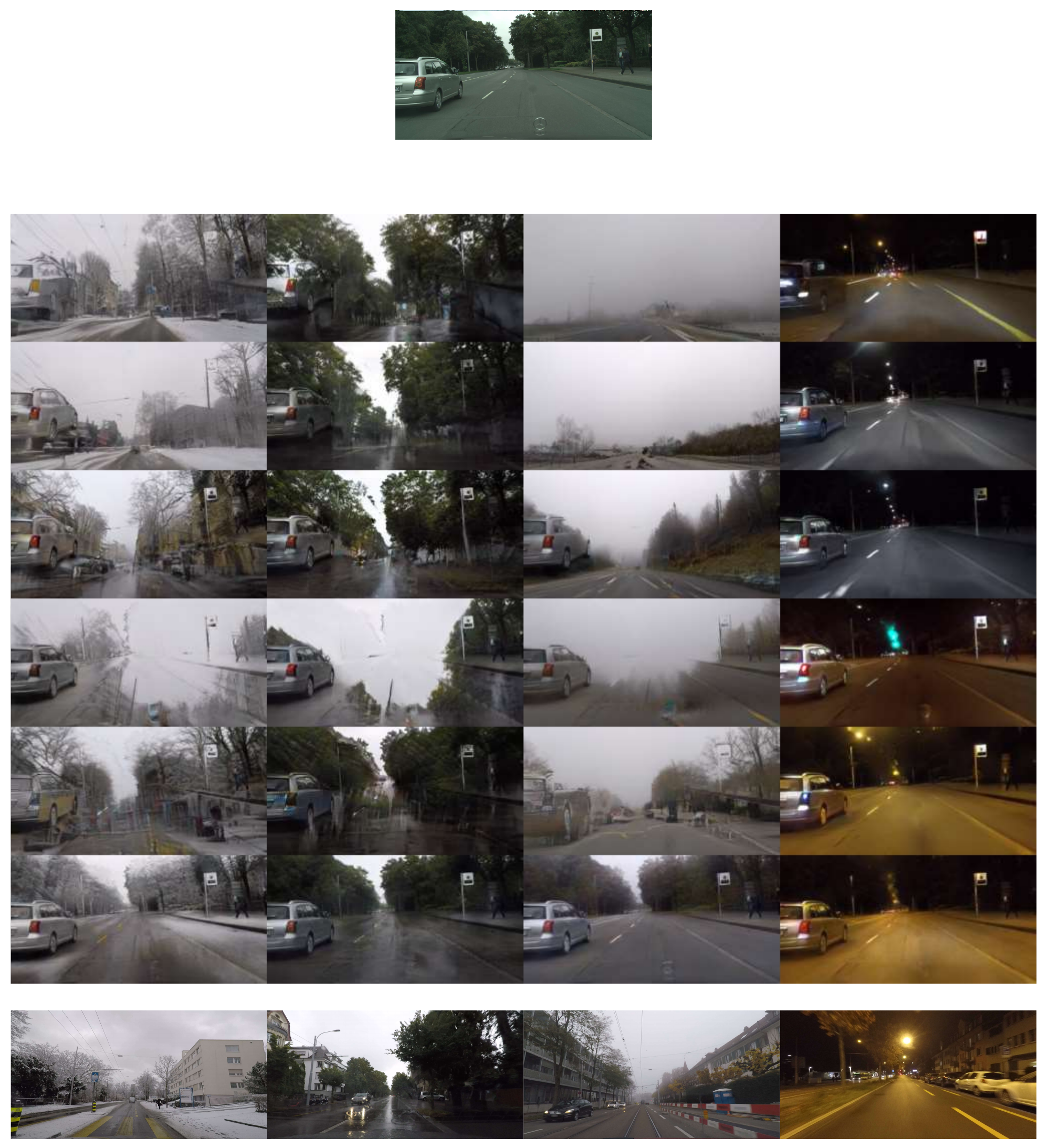}
	\vspace{-16.15cm}\\
  	\vspace{0cm}
	\raggedright{\scriptsize\hspace{180pt} Input}
	\vspace{2.65cm}\\
	\raggedright{\scriptsize\hspace{4pt} CycleGAN}
	\vspace{1.5cm}\\
	\raggedright{\scriptsize\hspace{12pt} UNIT}
	\vspace{1.5cm}\\
	\raggedright{\scriptsize\hspace{8pt}	MUNIT}
	\vspace{1.5cm}\\
	\raggedright{\scriptsize\hspace{12pt}	TSIT}
	\vspace{1.5cm}\\
	\raggedright{\scriptsize\hspace{8pt}	MGUIT}
	\vspace{1.5cm}\\
	\raggedright{\scriptsize\hspace{4pt} SHUNIT (ours)}
	\vspace{1.85cm}\\
	\raggedright{\scriptsize\hspace{6pt} Target domain}
	\vspace{1cm}
	\caption{\textbf{More qualitative comparison on Cityscapes (clear) $\rightarrow$ ACDC (snow/rain/fog/night).} From the given clear image (first row), we generate four adverse condition images using \cite{zhu2017unpaired, liu2017unsupervised, huang2018multimodal, jiang2020tsit, jeong2021memory}
	and SHUNIT (top row to bottom row order). In the last row, we show a sample of the real image for each adverse condition.
	}
	\label{fig:c2a_1}
\end{figure*}

\begin{figure*}[t]
	\hfill
	\includegraphics[width=.87\linewidth]{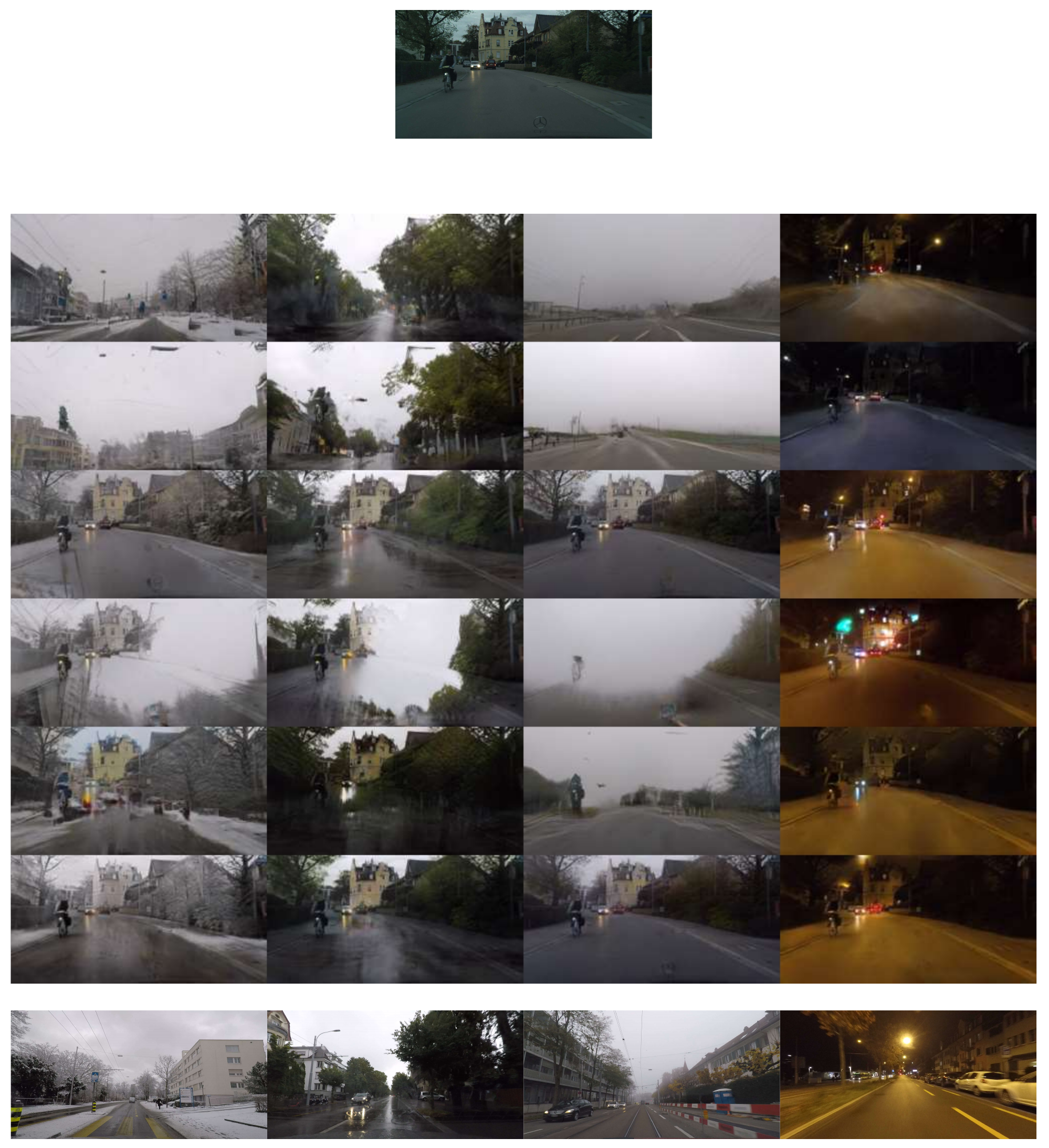}
	\vspace{-16.15cm}\\
  	\vspace{0cm}
	\raggedright{\scriptsize\hspace{180pt} Input}
	\vspace{2.65cm}\\
	\raggedright{\scriptsize\hspace{4pt} CycleGAN}
	\vspace{1.5cm}\\
	\raggedright{\scriptsize\hspace{12pt} UNIT}
	\vspace{1.5cm}\\
	\raggedright{\scriptsize\hspace{8pt}	MUNIT}
	\vspace{1.5cm}\\
	\raggedright{\scriptsize\hspace{12pt}	TSIT}
	\vspace{1.5cm}\\
	\raggedright{\scriptsize\hspace{8pt}	MGUIT}
	\vspace{1.5cm}\\
	\raggedright{\scriptsize\hspace{4pt} SHUNIT (ours)}
	\vspace{1.85cm}\\
	\raggedright{\scriptsize\hspace{6pt} Target domain}
	\vspace{1cm}
	\caption{\textbf{More qualitative comparison on Cityscapes (clear) $\rightarrow$ ACDC (snow/rain/fog/night).} From the given clear image (first row), we generate four adverse condition images using \cite{zhu2017unpaired, liu2017unsupervised, huang2018multimodal, jiang2020tsit, jeong2021memory}
	and SHUNIT (top row to bottom row order). In the last row, we show a sample of the real image for each adverse condition.
	}
	\label{fig:c2a_2}
\end{figure*}

\begin{figure*}[t]
	\hfill
	\includegraphics[width=.87\linewidth]{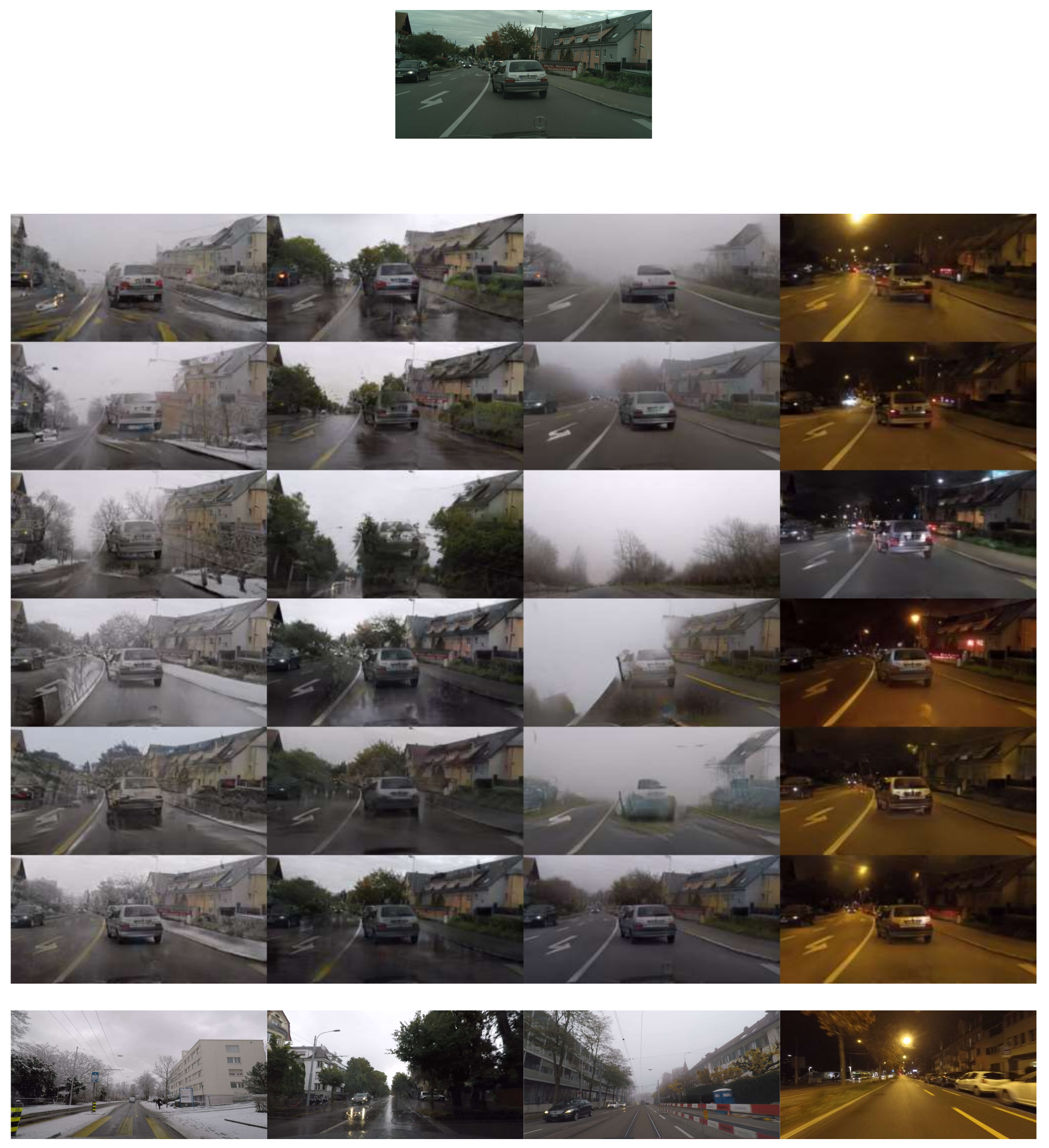}
	\vspace{-16.15cm}\\
  	\vspace{0cm}
	\raggedright{\scriptsize\hspace{180pt} Input}
	\vspace{2.65cm}\\
	\raggedright{\scriptsize\hspace{4pt} CycleGAN}
	\vspace{1.5cm}\\
	\raggedright{\scriptsize\hspace{12pt} UNIT}
	\vspace{1.5cm}\\
	\raggedright{\scriptsize\hspace{8pt}	MUNIT}
	\vspace{1.5cm}\\
	\raggedright{\scriptsize\hspace{12pt}	TSIT}
	\vspace{1.5cm}\\
	\raggedright{\scriptsize\hspace{8pt}	MGUIT}
	\vspace{1.5cm}\\
	\raggedright{\scriptsize\hspace{4pt} SHUNIT (ours)}
	\vspace{1.85cm}\\
	\raggedright{\scriptsize\hspace{6pt} Target domain}
	\vspace{1cm}
	\caption{\textbf{More qualitative comparison on Cityscapes (clear) $\rightarrow$ ACDC (snow/rain/fog/night).} From the given clear image (first row), we generate four adverse condition images using \cite{zhu2017unpaired, liu2017unsupervised, huang2018multimodal, jiang2020tsit, jeong2021memory}
	and SHUNIT (top row to bottom row order). In the last row, we show a sample of the real image for each adverse condition.
	}
	\label{fig:c2a_3}
\end{figure*}

\begin{figure*}[t]
	\hfill
	\includegraphics[width=.87\linewidth]{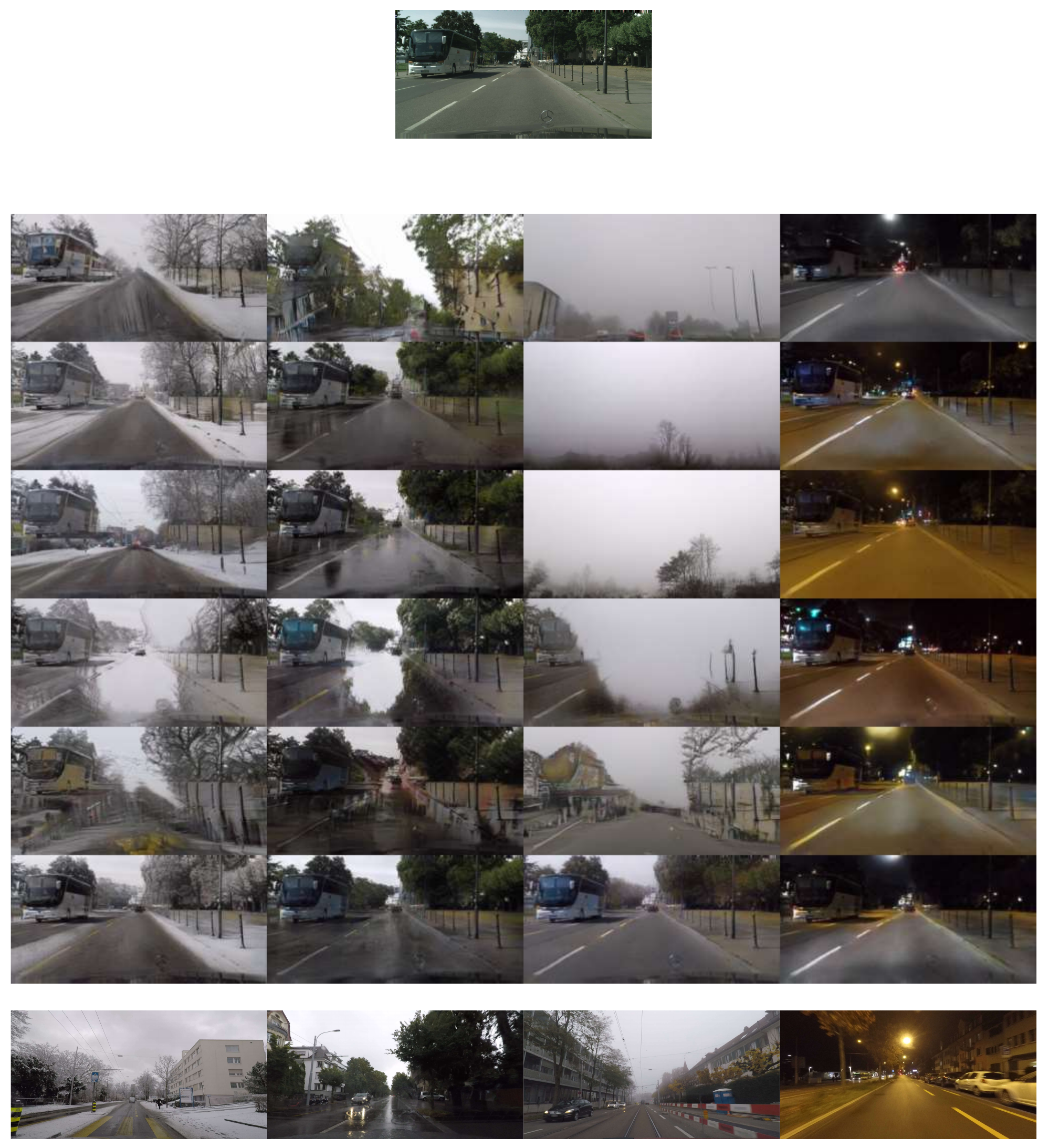}
	\vspace{-16.15cm}\\
  	\vspace{0cm}
	\raggedright{\scriptsize\hspace{180pt} Input}
	\vspace{2.65cm}\\
	\raggedright{\scriptsize\hspace{4pt} CycleGAN}
	\vspace{1.5cm}\\
	\raggedright{\scriptsize\hspace{12pt} UNIT}
	\vspace{1.5cm}\\
	\raggedright{\scriptsize\hspace{8pt}	MUNIT}
	\vspace{1.5cm}\\
	\raggedright{\scriptsize\hspace{12pt}	TSIT}
	\vspace{1.5cm}\\
	\raggedright{\scriptsize\hspace{8pt}	MGUIT}
	\vspace{1.5cm}\\
	\raggedright{\scriptsize\hspace{4pt} SHUNIT (ours)}
	\vspace{1.85cm}\\
	\raggedright{\scriptsize\hspace{6pt} Target domain}
	\vspace{1cm}
	\caption{\textbf{More qualitative comparison on Cityscapes (clear) $\rightarrow$ ACDC (snow/rain/fog/night).} From the given clear image (first row), we generate four adverse condition images using \cite{zhu2017unpaired, liu2017unsupervised, huang2018multimodal, jiang2020tsit, jeong2021memory}
	and SHUNIT (top row to bottom row order). In the last row, we show a sample of the real image for each adverse condition.
	}
	\label{fig:c2a_4}
\end{figure*}

We show more qualitative results on Cityscapes (clear) $\rightarrow$ ACDC (snow/rain/fog/night) in \Cref{fig:c2a_1,fig:c2a_2,fig:c2a_3,fig:c2a_4}.
In the results, we additionally show the results of UNIT~\cite{liu2017unsupervised} and TSIT~\cite{jiang2020tsit}, which are omitted in the main paper.
The qualitative results demonstrate that our method generate more realistic images than the previous works~\cite{zhu2017unpaired, liu2017unsupervised, huang2018multimodal, jiang2020tsit, jeong2021memory} in most cases. We further provide a comparison video in the supplementary material.

\subsection{KITTI $\rightarrow$ Cityscapes}
\begin{figure*}[t!]
 	\centering
	\includegraphics[width=\linewidth]{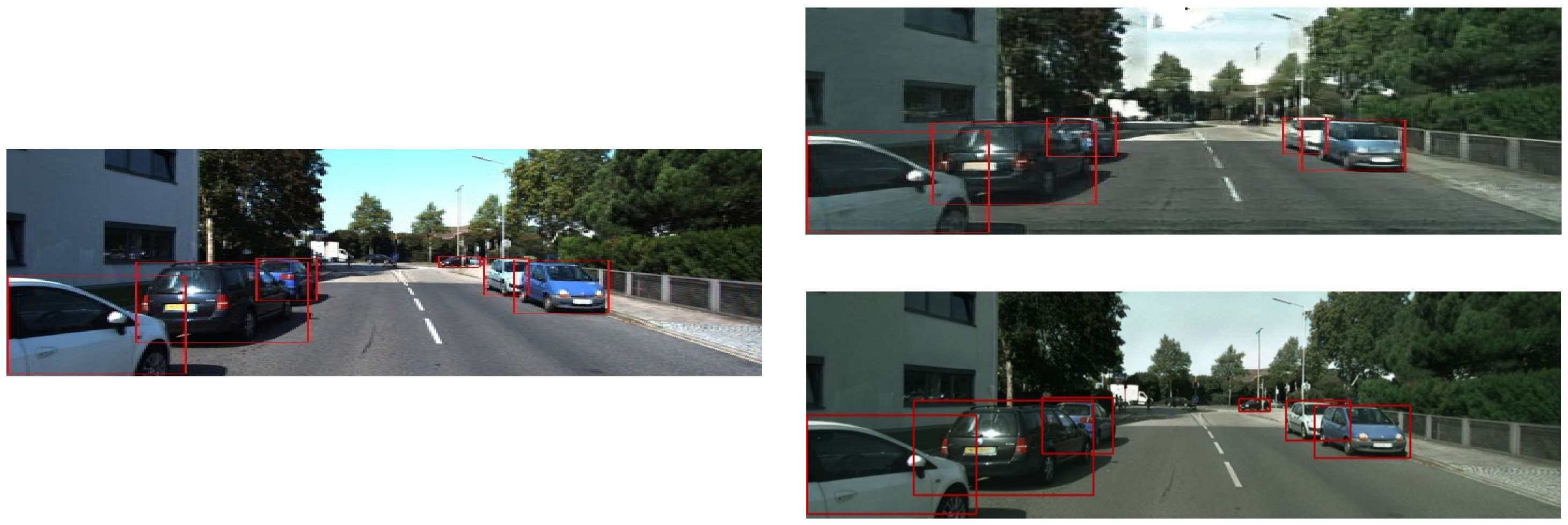}
	\vspace{-1.85cm}\\
	\vspace{-1.95cm}
	\raggedright{\small\hspace{360pt} (b) InstaFormer}
	\vspace{1.3cm}\\
	\raggedright{\small\hspace{90pt} (a) Input image (KITTI)}
	\vspace{1.15cm}\\
	\raggedright{\small\hspace{360pt} (c) SHUNIT (ours)}
	\vspace{1.0cm}\\
	\vspace{-1.0cm}
	\caption{\textbf{Qualitative comparison on domain adaptive detection for KITTI $\rightarrow$ Cityscapes.} Given (a) input image (KITTI), (b) and (c) show the translated image and the object detection result generated by InstaFormer and our method, respectively. Note that the results are taken from InstaFormer paper.}
	\label{fig:da}
\end{figure*}

Fig.~\ref{fig:da} shows the visual comparison of InstaFormer~\cite{kim2022instaformer} and our method for domain adaptive detection on KITTI $\rightarrow$ Cityscapes.
As shown in the figure, InstaFormer generates artifacts in the sky and struggles to maintain the sub-object components of a small object, such a car located in the center. In contrast, our method translates the sky to the Cityscapes style without artifacts, and preserves the sub-objects components of a small object. The more realistic image translation is demonstrated by a detection result: the Faster-RCNN~\cite{ren2015faster} detects the small car located in the center in our image (Fig.~\ref{fig:da}c) while cannot detect it in InstaFormer image (Fig.~\ref{fig:da}b)

\section{More Quantitative Results} \label{sec:4}
\Cref{table:mIoU_snow,table:mIoU_rain,table:mIoU_fog,table:mIoU_night} provide the per-class IoU performance on ACDC test set.
As shown in the tables, we achieve the best performance not only in mIoU but also in per-class IoU of the most classes on snow, rain, and fog.

\begin{table*}[t]
\centering
\resizebox{\textwidth}{!}
{
\setlength\tabcolsep{1pt}
\footnotesize
\begin{tabular}[b]{l|ccccccccccccccccccc|c}\toprule\noalign{\smallskip}
Methods            & {\rotatebox{90}{road }}          & {\rotatebox{90}{sidewalk }}      & {\rotatebox{90}{building }}      & {\rotatebox{90}{wall }}          & {\rotatebox{90}{fence }}         & {\rotatebox{90}{pole }}          & {\rotatebox{90}{traffic light }} & {\rotatebox{90}{traffic sign }}  & {\rotatebox{90}{vegetation }}    & {\rotatebox{90}{terrain }}       & {\rotatebox{90}{sky }}           & {\rotatebox{90}{person }}        & {\rotatebox{90}{rider }}         & {\rotatebox{90}{car }}           & {\rotatebox{90}{truck }}         & {\rotatebox{90}{bus }}           & {\rotatebox{90}{train }}        & {\rotatebox{90}{motorcycle }}    & {\rotatebox{90}{bicycle }}       & mIoU          \\\cmidrule{1-21}\morecmidrules \cmidrule{1-21}
CycleGAN~\cite{zhu2017unpaired}     & 54.55 & 12.74  & 41.24  & 8.42  & 9.71  & 11.78 & 23.77 & 28.34 & 38.86  & 1.68    & 2.07  & 17.85  & 17.20  & 62.00    & 6.05  & 34.29 & 50.93 & 8.94    & 19.57   & 23.68          \\
UNIT~\cite{liu2017unsupervised}         & 58.88 & 16.27  & 43.10   & 9.74  & 14.35 & 17.47 & 38.47 & 39.71 & 47.97  & 6.98    & 4.17  & 41.25  & 30.11 & 73.35 & 20.49 & 36.20  & 60.74 & 6.75    & 27.48   & 31.24          \\
MUNIT~\cite{huang2018multimodal}        & 65.09 & 21.06  & 44.42  & 13.53 & 18.20  & 19.04 & 41.24 & 40.68 & 51.58  & 7.94    & 5.85  & 41.21  & 27.10  & 77.44 & \textbf{24.08} & 31.01 & 60.19 & 18.93   & 34.14   & 33.83          \\
TSIT~\cite{jiang2020tsit}         & 79.32 & 42.91  & 47.03  & 13.01 & 18.48 & 21.82 & 45.05 & 42.71 & 55.98  & \textbf{10.05}   & 7.55  & 47.79  & \textbf{39.11} & \textbf{78.97} & 19.21 & \textbf{49.75} & 58.25 & 19.55   & 27.07   & 38.08          \\
MGUIT~\cite{jeong2021memory}        & 69.14 & 31.58  & 49.37  & 7.13  & 6.84  & 18.31 & 28.06 & 32.98 & 64.92  & 9.20     & 51.08 & 32.05  & 37.06 & 74.26 & 4.57  & 30.31 & 60.42 & 6.81    & 19.15   & 33.33          \\
SHUNIT(ours) & \textbf{81.62} & \textbf{46.12}  & \textbf{52.33}  & \textbf{16.03} & \textbf{27.34} & \textbf{25.07} & \textbf{49.22} & \textbf{43.19} & \textbf{80.08}  & 9.41    & \textbf{54.61} & \textbf{52.95}  & 36.23 & 77.66 & 20.22 & 48.18 & \textbf{69.04} & \textbf{26.47}   & \textbf{43.12}   & \textbf{45.15} \\ \bottomrule
\end{tabular}
}
\caption{\label{table:mIoU_snow}\textbf{Per-class IoU results on Cityscapes (clear) $\rightarrow$ ACDC (snow).}}
\end{table*}

\begin{table*}[t]
\centering
\resizebox{\textwidth}{!}
{
\setlength\tabcolsep{1pt}
\footnotesize
\begin{tabular}[b]{l|ccccccccccccccccccc|c}\toprule\noalign{\smallskip}
Methods            & {\rotatebox{90}{road }}          & {\rotatebox{90}{sidewalk }}      & {\rotatebox{90}{building }}      & {\rotatebox{90}{wall }}          & {\rotatebox{90}{fence }}         & {\rotatebox{90}{pole }}          & {\rotatebox{90}{traffic light }} & {\rotatebox{90}{traffic sign }}  & {\rotatebox{90}{vegetation }}    & {\rotatebox{90}{terrain }}       & {\rotatebox{90}{sky }}           & {\rotatebox{90}{person }}        & {\rotatebox{90}{rider }}         & {\rotatebox{90}{car }}           & {\rotatebox{90}{truck }}         & {\rotatebox{90}{bus }}           & {\rotatebox{90}{train }}        & {\rotatebox{90}{motorcycle }}    & {\rotatebox{90}{bicycle }}       & mIoU          \\\cmidrule{1-21}\morecmidrules \cmidrule{1-21}
CycleGAN~\cite{zhu2017unpaired}     & 73.51 & 31.73  & 51.75  & 20.75 & 18.66 & 20.99 & 24.80  & 26.39 & 71.58  & 24.46   & 28.91 & 34.06  & 11.75 & 74.66 & 37.32 & 23.02  & 52.42 & 16.18   & 40.19   & 35.96          \\
UNIT~\cite{liu2017unsupervised}         & 74.20  & 37.16  & 58.67  & 24.76 & 17.98 & 23.51 & 38.65 & 44.07 & 58.10   & 20.11   & 18.63 & 40.51  & 15.17 & 78.89 & 41.53 & \textbf{40.16}  & 57.62 & 18.57   & 40.20    & 39.39          \\
MUNIT~\cite{huang2018multimodal}        & 77.68 & 40.79  & 71.64  & 20.81 & 27.55 & 26.40  & 44.48 & 43.77 & 67.15  & 24.84   & 55.84 & 42.86  & 15.69 & 79.60  & 41.18 & 38.71 & 57.21 & 20.00      & 43.53   & 44.20           \\
TSIT~\cite{jiang2020tsit}         & \textbf{78.08} & \textbf{46.43}  & 71.28  & \textbf{29.44} & 25.12 & \textbf{29.23} & 44.95 & 46.64 & 78.14  & 24.68   & 67.75 & \textbf{44.24}  & \textbf{16.37} & \textbf{80.94} & 46.28 & 31.09  & 57.90  & 25.87   & 41.74   & 46.40           \\
MGUIT~\cite{jeong2021memory}        & 73.69 & 31.94  & 80.06  & 19.54 & 14.44 & 24.74 & 37.17 & 34.91 & 79.99  & 21.61   & 93.74 & 34.46  & 13.60  & 76.69 & 27.54 & 38.64  & \textbf{60.43} & 16.32   & 31.04   & 42.60           \\
SHUNIT(ours) & 76.41 & 28.42  & \textbf{82.03}  & 23.44 & \textbf{28.12} & 28.81 & \textbf{46.15} & \textbf{46.78} & \textbf{86.54}  & \textbf{36.26}   & \textbf{94.36} & 41.39  & 12.79 & 80.74 & \textbf{46.90}  & 25.71  & 58.57 & \textbf{29.93}   & \textbf{45.09}   & \textbf{48.84} \\ \bottomrule
\end{tabular}
}
\caption{\label{table:mIoU_rain}\textbf{Per-class IoU results on Cityscapes (clear) $\rightarrow$ ACDC (rain).}}
\end{table*}

\begin{table*}[h]
\centering
\resizebox{\textwidth}{!}
{
\setlength\tabcolsep{1pt}
\footnotesize
\begin{tabular}[b]{l|ccccccccccccccccccc|c}\toprule\noalign{\smallskip}
Methods            & {\rotatebox{90}{road }}          & {\rotatebox{90}{sidewalk }}      & {\rotatebox{90}{building }}      & {\rotatebox{90}{wall }}          & {\rotatebox{90}{fence }}         & {\rotatebox{90}{pole }}          & {\rotatebox{90}{traffic light }} & {\rotatebox{90}{traffic sign }}  & {\rotatebox{90}{vegetation }}    & {\rotatebox{90}{terrain }}       & {\rotatebox{90}{sky }}           & {\rotatebox{90}{person }}        & {\rotatebox{90}{rider }}         & {\rotatebox{90}{car }}           & {\rotatebox{90}{truck }}         & {\rotatebox{90}{bus }}           & {\rotatebox{90}{train }}        & {\rotatebox{90}{motorcycle }}    & {\rotatebox{90}{bicycle }}       & mIoU          \\\cmidrule{1-21}\morecmidrules \cmidrule{1-21}
CycleGAN~\cite{zhu2017unpaired}     & 46.89 & 35.49  & 33.79  & \textbf{27.63} & \textbf{27.07} & \textbf{17.65} & 14.74 & 10.70  & 7.82   & 7.14    & 6.96  & 5.39   & 5.28  & 3.81  & 3.08  & 2.57  & 2.28  & 2.25    & 0.32    & 13.73          \\
UNIT~\cite{liu2017unsupervised}         & 64.21 & 26.65  & 30.18  & 16.54 & 11.36 & 10.77 & 23.89 & 30.93 & 31.44  & 25.16   & 1.68  & 14.22  & 4.55  & 58.45 & 34.28 & 45.42 & 41.64 & 27.20    & 5.79    & 28.70           \\
MUNIT~\cite{huang2018multimodal}        & 54.73 & 26.48  & 28.43  & 14.60  & 8.54  & 7.36  & 30.30  & 29.62 & 33.86  & 24.07   & 3.95  & 8.33   & 32.37 & 56.23 & 30.05 & \textbf{58.80}  & \textbf{51.89} & 12.87   & 8.93    & 27.44          \\
TSIT~\cite{jiang2020tsit}         & \textbf{80.10}  & \textbf{46.44}  & 19.39  & 14.77 & 15.83 & 16.11 & 27.27 & \textbf{39.82} & \textbf{72.30}   & \textbf{44.35}   & 2.44  & 25.78  & 41.78 & \textbf{65.67} & \textbf{41.98} & 49.48 & 42.54 & \textbf{31.12}   & \textbf{19.71}   & 36.68          \\
MGUIT~\cite{jeong2021memory}        & 54.26 & 11.66  & 21.85  & 5.81  & 0.89  & 8.73  & 3.86  & 11.28 & 22.51  & 7.30     & 1.37  & 1.97   & 3.08  & 34.31 & 1.08  & 0.05  & 3.86  & 0.00       & 0.34    & 10.22          \\
SHUNIT(ours) & 72.35 & 43.36  & \textbf{41.68}  & 21.42 & 18.13 & 16.08 & \textbf{39.11} & 35.90  & 71.56  & 35.10    & \textbf{67.93} & \textbf{27.53}  & \textbf{50.45} & 52.55 & 37.31 & 37.06 & 28.32 & 27.64   & 16.57   & \textbf{38.96} \\ \bottomrule
\end{tabular}
}
\caption{\label{table:mIoU_fog}\textbf{Per-class IoU results on Cityscapes (clear) $\rightarrow$ ACDC (fog).}}
\end{table*}

\begin{table*}[h]
\centering
\resizebox{\textwidth}{!}
{
\setlength\tabcolsep{1pt}
\footnotesize
\begin{tabular}[b]{l|ccccccccccccccccccc|c}\toprule\noalign{\smallskip}
Methods            & {\rotatebox{90}{road }}          & {\rotatebox{90}{sidewalk }}      & {\rotatebox{90}{building }}      & {\rotatebox{90}{wall }}          & {\rotatebox{90}{fence }}         & {\rotatebox{90}{pole }}          & {\rotatebox{90}{traffic light }} & {\rotatebox{90}{traffic sign }}  & {\rotatebox{90}{vegetation }}    & {\rotatebox{90}{terrain }}       & {\rotatebox{90}{sky }}           & {\rotatebox{90}{person }}        & {\rotatebox{90}{rider }}         & {\rotatebox{90}{car }}           & {\rotatebox{90}{truck }}         & {\rotatebox{90}{bus }}           & {\rotatebox{90}{train }}        & {\rotatebox{90}{motorcycle }}    & {\rotatebox{90}{bicycle }}       & mIoU          \\\cmidrule{1-21}\morecmidrules \cmidrule{1-21}
CycleGAN~\cite{zhu2017unpaired}     & 86.03 & 46.64  & 60.57  & 24.69 & 15.16 & 26.56 & 11.75 & 31.92 & 42.44  & 27.05   & 4.47  & 36.84  & 10.44 & 61.32 & 0.04  & 15.97 & \textbf{51.54} & 16.27   & 25.51   & 31.33          \\
UNIT~\cite{liu2017unsupervised}         & 86.03 & 46.74  & 63.33  & 26.69 & 14.80  & 28.41 & 19.25 & 30.92 & 43.94  & 35.14   & 15.34 & 40.08  & 19.55 & 64.43 & 2.51  & \textbf{30.81} & 44.15 & \textbf{29.44}   & 29.00      & 35.29          \\
MUNIT~\cite{huang2018multimodal}        & 87.73 & \textbf{68.58}  & 66.92  & \textbf{54.84} & \textbf{51.74} & \textbf{44.49} & \textbf{42.45} & \textbf{35.94} & 33.97  & 32.46   & \textbf{30.67} & 29.12  & \textbf{27.61} & 26.24 & \textbf{20.50}  & 19.91 & 18.52 & 16.51   & 3.01    & \textbf{37.43} \\
TSIT~\cite{jiang2020tsit}         & \textbf{88.69} & 56.81  & 63.49  & 21.67 & 16.59 & 27.28 & 21.60  & 33.81 & 43.53  & \textbf{38.21}   & 3.50   & \textbf{42.42}  & 19.94 & \textbf{67.13} & 4.08  & 30.27 & 45.36 & 25.55   & \textbf{32.62}   & 35.92          \\
MGUIT~\cite{jeong2021memory}        & 86.67 & 44.75  & 66.36  & 24.50  & 16.13 & 28.26 & 8.87  & 32.15 & 42.24  & 8.32    & 8.66  & 37.88  & 10.85 & 61.65 & 2.38  & 19.17 & 46.22 & 20.52   & 30.16   & 31.36          \\
SHUNIT(ours) & 85.65 & 44.84  & \textbf{67.98}  & 24.53 & 18.98 & 26.66 & 20.57 & 30.32 & \textbf{45.02}  & 26.21   & 16.61 & 39.13  & 14.40  & 62.68 & 0.56  & 21.13 & 42.92 & 23.59   & 27.76   & 33.66 \\ \bottomrule
\end{tabular}
}
\caption{\label{table:mIoU_night}\textbf{Per-class IoU results on Cityscapes (clear) $\rightarrow$ ACDC (night).}}
\end{table*}

\newpage
\qquad
\clearpage

\end{document}